\documentclass[preprint,10pt,onehalfspacing]{elsarticle}
\usepackage{lineno}

\usepackage{url,hyperref,lineno,microtype,subcaption}
\usepackage[onehalfspacing]{setspace}

\usepackage[whole]{bxcjkjatype} 

\usepackage{amsmath,amsthm,ascmac,graphicx,color}
\usepackage{amssymb}  
\usepackage{amsfonts}
\usepackage{natbib}

\usepackage{algorithm}
\usepackage{algpseudocode}

\theoremstyle{plain}

\newtheorem*{thm*}{Theorem}
\theoremstyle{remark} 

\usepackage{diagbox}
\usepackage{enumerate}
\usepackage{multirow}
\usepackage{multibib}

\usepackage[normalem]{ulem}
\theoremstyle{plain}

\theoremstyle{definition}

\theoremstyle{remark}

\usepackage{mdframed} 

\usepackage[whole]{bxcjkjatype} 

\renewcommand{\emph}[1]{\textit{#1}}  

\begin{document}

\begin{frontmatter}

\title{Generative Emergent Communication:\\ Large Language Model is a Collective World Model}

\author[1]{Tadahiro Taniguchi\corref{cor1}}
\ead{taniguchi@i.kyoto-u.ac.jp} 

\author[2]{Ryo Ueda}

\author[3]{Tomoaki Nakamura}

\author[2]{Masahiro Suzuki}

\author[4]{Akira Taniguchi}

\cortext[cor1]{Corresponding author}

\affiliation[1]{organization={Graduate School of Informatics, Kyoto University},
                country={Japan}}
\affiliation[2]{organization={Graduate School of Engineering, The University of Tokyo},
                country={Japan}}
\affiliation[3]{organization={Graduate School of Informatics and Engineering, The University of Electro-Communications},
                country={Japan}}
\affiliation[4]{organization={College of Information Science and Engineering, Ritsumeikan University},
                country={Japan}}

\begin{abstract}
Large Language Models (LLMs) have demonstrated a remarkable ability to capture extensive world knowledge, yet how this is achieved without direct sensorimotor experience remains a fundamental puzzle. This study proposes a novel theoretical solution by introducing the \emph{Collective World Model} hypothesis. We argue that an LLM does not learn a world model from scratch; instead, it learns a statistical approximation of a collective world model that is already implicitly encoded in human language through a society-wide process of embodied, interactive sense-making.
To formalize this process, we introduce \emph{generative emergent communication} (Generative EmCom), a framework built on the Collective Predictive Coding (CPC). This framework models the emergence of language as a process of decentralized Bayesian inference over the internal states of multiple agents. We argue that this process effectively creates an encoder-decoder structure at a societal scale: human society collectively encodes its grounded, internal representations into language, and an LLM subsequently decodes these symbols to reconstruct a latent space that mirrors the structure of the original collective representations.
This perspective provides a principled, mathematical explanation for how LLMs acquire their capabilities. The main contributions of this paper are: 1) the formalization of the Generative EmCom framework, clarifying its connection to world models and multi-agent reinforcement learning, and 2) its application to interpret LLMs, explaining phenomena such as distributional semantics as a natural consequence of representation reconstruction. This work provides a unified theory that bridges individual cognitive development, collective language evolution, and the foundations of large-scale AI.
\end{abstract}


\begin{keyword}
emergent communication \sep large language model \sep world model \sep generative model \sep symbol emergence
\end{keyword}

\end{frontmatter}

\section{Introduction}

Large Language Models (LLMs) have achieved astonishing success, exhibiting a profound capacity for reasoning and knowledge retrieval across countless domains~\citep{brown2020language,Min2023-dp}. A central puzzle, however, lies in their apparent grasp of the structure of the physical world. LLMs, trained on vast corpora of text, are not designed to interact with an environment and lack any direct sensorimotor experience~\citep{andreas2022language,bender2021dangers}. Yet, they seem to possess what could be described as a ``world model.'' How can an LLM acquire such a rich model of the world without ever having perceived or acted within it? This paper proposes a theoretical solution to this fundamental question.


The debate over whether LLMs truly possess world models is active and ongoing. A growing body of evidence suggests that LLMs acquire surprisingly rich representations of the real world. For instance, some studies show that LLMs implicitly learn representations corresponding to vision and hearing just by reading text \citep{verma2025implicitly}, and that specific spatiotemporal representations emerge within their activations \citep{gurnee2023language}. Furthermore, this internal knowledge proves functionally potent; LLMs can be used as world models for complex planning tasks \citep{hao2023reasoning}, and their distributional semantics can be grounded in the physical world to guide robotic perception and action \citep{osada2024reflectance, yoshida2025text}. Conversely, other work has revealed that LLMs are not reliable world simulators, suggesting their internal models are brittle or fundamentally different from those of embodied agents \citep{wang2024can}. This conflicting evidence highlights a deep conceptual gap in our understanding.

To clarify our argument, we distinguish between two concepts of a world model~\citep{taniguchi2023world,ding2024understanding}:

\begin{enumerate}[(1)]
    \item \textbf{Type 1 World Model (Internal Model)}: A subjective, internal model that an agent learns through its own sensorimotor interactions with the environment to predict future states and plan actions. This corresponds to the agent's \emph{Umwelt}~\citep{Uexkull}.
    \item \textbf{Type 2 World Model (Model of the World)}: An objective, structured representation of knowledge about the world, its entities, and their relations, which may not be tied to a single agent's direct experience.
\end{enumerate}

We can frame this distinction as one between a subjective \emph{Type 1 World Model} and an objective \emph{Type 2 World Model}. Much of the confusion in current discourse arises from this distinction; while the term "world model" in AI often refers to a Type 1 model, particularly in the context of model-based reinforcement learning and predictive coding~\citep{ha2018world,hafner2019learning,hafner2019dream,friston2021world,taniguchi2023world}, discussions about LLMs often imply they possess a Type 2 model. This presents a paradox: it is difficult to see how a disembodied LLM could form a subjective, embodied Type 1 model, yet it is equally unclear how it could acquire an objective Type 2 model without any access to the world.

This paper proposes a solution to this paradox. We argue that the seemingly objective knowledge within an LLM is, in fact, a pseudo-objective structure encoded in the distributional semantics of language. This structure emerges as a result of aggregating countless subjective, Type 1 world models from a society of embodied agents. From the perspective of \emph{generative emergent communication} (Generative EmCom) an LLM does not model the world directly, but rather models the \emph{collective} of these Type 1 models as externalized in language. This leads to our central hypothesis:

\begin{mdframed}[linewidth=0.5pt, roundcorner=5pt, innertopmargin=5pt, innerbottommargin=5pt]
\textbf{The Collective World Model Hypothesis}

Human language is not merely a communication protocol but serves as an externalized representation of a \textbf{collective world model}, which emerges from the decentralized, interactive sense-making processes of an entire society of embodied agents. LLMs acquire their world knowledge by learning a statistical approximation of this collective world model encoded in text corpora.
\end{mdframed}

To provide a formal basis for this hypothesis, we introduce a new theoretical framework called \emph{Generative EmCom}. This framework is built upon the \emph{Collective Predictive Coding} (CPC)~\citep{taniguchi2024collective}, which extends the principles of predictive coding, the free-energy principle (FEP) and the Bayesian brain hypothesis~\citep{hohwy2013predictive,friston2010free,friston2019free,Doya2007} from individual brains to a societal level. Generative EmCom formalizes how a shared symbol system, i.e., language, emerges as multiple agents collectively seek to minimize their shared prediction errors about the world, a process we model as decentralized Bayesian inference.

However, such an integrative theory has been largely missing. The series of studies on emergent communication (EmCom) and symbol emergence has attempted to explain the formation of language~\citep{peters2024survey,LazaridouB-2020-emergent}. Yet, these approaches have often failed to bridge the gap between two interdependent aspects: first, the learning of an individual agent's world model, which is grounded in its embodiment and environmental adaptation~\citep{friston2021world,taniguchi2023world}; and second, the collective emergence of a language that reflects this grounded knowledge. Our work aims to address this specific challenge by providing a unified framework.

This study, therefore, makes two primary contributions. First, we formalize the Generative EmCom framework, clarifying its relationship with conventional approaches and demonstrating its utility in multi-agent systems. Second, using this framework, we provide a principled, mathematical interpretation of LLMs as collective world models, offering a coherent explanation for their otherwise mysterious capabilities. This unified perspective bridges the gap between EmCom, cognitive development, and the foundations of large-scale AI, opening new frontiers for research. Indeed, the theoretical framework proposed herein has already begun to inspire new concrete models for multi-agent coordination in dynamic environments~\citep{nomura2025decentralized} and reward-independent communication in multi-agent reinforcement learning (MARL)~\citep{yoshida2025reward}.

The remainder of this paper is organized as follows: Section 2 reviews the current landscape of EmCom and world models, identifying the theoretical gap our work addresses. Section 3 details the theoretical framework of Generative EmCom. Section 4 explains how a collective world model can emerge from multi-agent interaction and cooperation. Section 5 applies this theory to reinterpret LLMs. Finally, Section 6 discusses the implications and limitations of our work, and Section 7 concludes the paper.

\section{The Landscape of EmCom and World Models}
Language evolves and changes over time as a result of decentralized human communications~\citep{deacon1998symbolic,peters2024survey,Steels97,Steels2007}. Sentences are generated to describe a wide range of phenomena, including external events, emotions, and intentions. In particular, the system of language is inherently dyanamics rather than static~\citep{wittgenstein2009philosophical,STEELS2011339,tomasello2005constructing,taniguchi2018symbol}. As Peirce, the founder of semiotics, suggested, symbols, including language, can be characterized by a triadic relationship of sign, object, and interpretant~\citep{peirce1974collected,peirce1991peirce,Chandler2002}. Here, sign corresponds to words, sentences and other signals. In particular, the correspondence between sign and object, which is signified by a sign (i.e., signifier), is determined by an interpretant. In other words, the meaning of a sign, that is, language, depends on culture and context, and so on.

\subsection{Conventional Approaches to EmCom and Their Limits}

The study of how communication protocols and symbolic language emerge from multi-agent interaction, a field known as EmCom, has been explored through several major approaches, including language games, MARL, and iterated learning models (ILMs)~\citep{peters2025emergent, LazaridouB-2020-emergent, STEELS2011339, zhu2024survey, brandizzi2023toward}. A central theme in this research, particularly in studies employing language games and powered by deep neural networks~\citep{FoersterAFW-2016-riddles, MordatchA-2018-emergence, JaquesLHGOSLF-2019-intrinsic-motivation}, is the extent to which the resulting protocols, or \emph{emergent languages} (EmLangs), resemble human language. For instance, questions have been raised about whether EmLangs exhibit compositionality~\citep{KotturMLB-2017-compositionality, chaabouni2020compositionality}, follow well-known statistical properties of natural languages~\citep{chaabouni2019anti-efficient, RitaCD-2020-lazimpa, UedaW-2021-ZLA, UedaIM-2023-HAS}, or can be shaped by cognitive constraints~\citep{RiUN-2023-attention, KatoUNM-2024-stack}.

A foundational paradigm for many of these investigations is the \emph{Lewis signaling game}~\citep{Lewis-1969-convention}. This approach, often aligned with Shannon's information theory~\citep{shannon1948mathematical}, can be regarded as a discriminative model for optimizing a communication protocol.

The signaling game is a simple communication model that involves only a sender \(S_{\boldsymbol{\phi}}(m|x)\) and a receiver \(R_{\boldsymbol{\theta}}(x|m)\) and only allows unidirectional communication from the sender to the receiver.
At each play, the signaling game proceeds as follows:
\begin{enumerate}
    \item \textbf{Observation}: Sender \(S_{\boldsymbol{\phi}}\) obtains an observation \(x\), that is, \(x\sim p(x)\).
    \item \textbf{Signaling}: Sender \(S_{\boldsymbol{\phi}}\) generates a message \(m\) from the observation \(x\), that is, \(m\sim S_{\boldsymbol{\phi}}(m|x)\).
    \item \textbf{Reconstruction}: Receiver \(R_{\boldsymbol{\theta}}\) attempts to reconstruct the original observation \(x\) from the message \(m\) via \(R_{\boldsymbol{\theta}}(x|m)\).
\end{enumerate}
Sender \(S_{\boldsymbol{\phi}}\) and receiver \(R_{\boldsymbol{\theta}}\) are optimized via a gradient-based method toward successful communication.
Conventionally, the objective function of the signaling game (to be maximized) is defined as follows \citep{ChaabouniKDB-2019-antiefficient, RitaTMGPDS-2022-emergent}:
\begin{align}
    \mathcal{J}_{\textup{MI}}(\boldsymbol{\phi}, \boldsymbol{\theta})
    \mathrel{:=}
    \mathbb{E}_{p(x),S_{\boldsymbol{\phi}}(m|x)}[\log R_{\boldsymbol{\theta}}(x|m)].
\end{align}
We refer to \(\mathcal{J}_{\textup{MI}}\) as the \emph{mutual information (MI)-maximizing objective function} because it is known to be an evidence lower bound (ELBO) (up to constant) of the following mutual information between \(X\) and \(M\) \citep{BarberA-2003-information, PooleOOAT-2019-on}:
\begin{align}
    I_{\boldsymbol{\phi}}(X;M)
    \mathrel{:=}
    \mathbb{E}_{p(x),S_{\boldsymbol{\phi}}(m|x)}\left[
        \log\frac{S_{\boldsymbol{\phi}}(m|x)}{\mathbb{E}_{p(x')}[S_{\boldsymbol{\phi}}(m|x')]}
    \right].
\end{align}
This implies that the conventional signaling game in the field of EmCom has been formulated as a problem of maximizing the mutual information between \(X\) and \(M\), where \( X \) and \( M \) denote random variables corresponding to the realizations \( x \) and \( m \), respectively.

However, from the viewpoint of our paper, it is more fruitful to reinterpret these communication games through the lens of probabilistic generative models. This shift from a discriminative to a generative perspective reveals deeper connections to principles of cognition, learning, and the information-theoretic foundations of frameworks like the Variational Autoencoder (VAE) and the Information Bottleneck (IB) principle~\citep{AlemiFDM-2017-VIB, ZaslavskyKRT-2018-efficient}.

This generative viewpoint reframes the signaling game as a problem of maximizing the ELBO~\citep{UedaT-2024-signaling-game-as-vae}. The objective function becomes:
\begin{align}
\begin{split}
    \mathcal{J}_{\textup{ELBO}}(\boldsymbol{\phi}, \boldsymbol{\theta})
    &\mathrel{:=}
    \mathbb{E}_{p(x)}[\mathbb{E}_{S_{\boldsymbol{\phi}}(m|x)}[\log R_{\boldsymbol{\theta}}(m|x)]
    \\
    &\hphantom{\mathrel{:=}\mathbb{E}_{p(x)}[}
    -\beta\textrm{KL}(S_{\boldsymbol{\phi}}(m|x)||p_{\boldsymbol{\theta}}(m))].
\end{split}
\end{align}
We refer to \(\mathcal{J}_{\textup{ELBO}}\) as the \emph{ELBO-maximizing objective function}, contrasting it with the MI-maximizing objective \(\mathcal{J}_{\textup{MI}}\).
By adopting the ELBO-maximizing objective function, we can introduce concepts from computational psycholinguistics into signaling games.
To observe this, let us transform the ELBO maximizing objective function as follows:
\begin{align}
\begin{split}
    \mathcal{J}_{\textup{ELBO}}(\boldsymbol{\phi}, \boldsymbol{\theta})
    &\mathrel{:=}
    \mathbb{E}_{p(x),S_{\boldsymbol{\phi}}(m|x)}[
        \underbrace{\log R_{\theta}(x|m)}_{\textup{communication}}
        +\beta\underbrace{\log p_{\theta}(m)}_{\substack{\textup{(negative)}\\\textup{surprisal}}}
    ]
    \\
    &\phantom{\mathrel{:=}}
    -\beta\underbrace{\mathbb{E}_{p(x)}\mathcal{H}(S_{\phi}(M|x))}_{\substack{\textup{entropy}\\\textup{maximizer}}},
\end{split}
\end{align}
where $\beta$ is a hyperparameter. Here, a term known as \emph{surprisal} appears, which is a concept commonly used in computational psycholinguistics \citep{Hale-2001-earley-parser, Levy-2008-expectation-based, KuribayashiOBI-2022-context}.
Surprisal is assumed to represent the cognitive load experienced by a listener/reader (or the receiver in the signaling game) when processing a sentence.
Therefore, the ELBO-maximizing signaling game naturally models the trade-off between information transmission and surprisal.

A similar discussion involves modeling the trade-off between information transmission and efficiency, where studies have used the (variational) information bottleneck (IB, VIB) framework to model communication \citep{ZaslavskyKRT-2018-efficient, ChaabouniKDB-2021-communicating, TuckerLSZ-2022-trading}.
In fact, VIB is proven to be a generalization of (beta-)VAE \citep{AlemiFDM-2017-VIB, AchilleS-2018-information-dropout}, resulting in similar models. In addition, a contemporary work \citep{ueda2024reinterpreting} shows that a variant framework known as a \emph{referential game} can also be reformulated with an ELBO-like objective, analogous in structure to that of a conditional VAE \citep{kingma2014cvae, sohn2015cvae}.

In relation to discussions in the field of evolutionary linguistics, some studies have also incorporated ILM into the EmCom framework \citep{RenGLCK-2020-compositional}, which is another important research theme.
The ILM is a framework that models generational changes, where supervised learning is repeatedly performed from parent agents to child agents.
In the context of VAE and VIB, however, little discussion exists on modeling generational changes, and this remains a future challenge when considering generative symbol emergence.

The formulation of the signaling game presented in this section, that is, generative model-based re-formulation of conventional EmCom, shares some fundamental connections with the generative frameworks discussed in later sections. All these formulations can be interpreted as representation learning with messages serving as latent variables in generative models when using ELBO-type (or VIB-type) objectives. However, these signaling game-based approaches have inherent limitations: they typically assume a simple two-agent setting with asymmetric sender-receiver roles, and their extension to populated, decentralized settings is not straightforward.

Beyond these foundational models, the field of Semantic Communication (SC) has gained traction, aiming to transmit information based on meaning and effectiveness rather than bit-level fidelity~\citep{popovski2020semantic}. This has spurred research into areas such as deep learning-based semantic encoders~\citep{xie2021deep}, contextual reasoning for shared understanding~\citep{seo2021semantics}, and learning goal-oriented languages from interaction~\citep{farshbafan2022common}. A comprehensive framework, Emergent Semantic Communication (ESC), seeks to integrate many of these threads by using causal reasoning in a neuro-symbolic architecture to create efficient emergent languages~\citep{thomas2023neuro}. While these approaches advance the engineering of goal-oriented communication, they do not address how a society-wide, general-purpose language emerges to reflect a \emph{collective} world model, which is the central question this paper addresses.

\subsection{World Models for Individual Cognitive Agents}

The concept of a world model represents an internal model within an agent that captures the dynamics of environmental states, their responses to the actions of the agent, and their relationships with sensory inputs~\citep{ha2018recurrent,friston2021world,taniguchi2023world,ding2024understanding}. The concept of {\it world model} has its origins in the early days of AI and robotics studies~\citep{nilsson1984shakey}. Initial studies on ML investigated techniques for agents to autonomously construct and adapt their world models~\citep{schmidhuber1990making, sutton1990integrated}. Currently, the term generally refers to predictive frameworks~\citep{hafner2019dream}, predominantly implemented using deep neural network architectures.

The concept of world models is closely related to the idea of predictive coding (PC). PC is the idea that the brain constantly predicts sensory information and updates its internal models to minimize prediction errors.
In the context of cognitive robotics and AI, world models provide the structure for representing and reasoning about the environment, whereas PC offers a mechanism for learning and updating these models based on sensory experiences. The free energy principle and active inference further unify these concepts, suggesting that both perception and action can be considered as processes of minimizing prediction errors or free energy. A theoretical connection exists between them~\citep{taniguchi2023world} .

Generally, a theory of world models can be based on partially observable Markov decision process (POMDP). In this framework, the state $z_t$ at time $t$ is not directly observable by the agent. Instead, the agent receives an observation $x_t$, which is assumed to be generated from a latent state $z_t$. The agent's actions $a_t$ influence the transition of states according to a probability distribution $p(z_{t+1}|z_t,a_t)$. The observation model is given by $p(x_t|z_t)$. The goal of the agent is to learn these probability distributions and use them to make predictions and inferences about the environment. This can be formalized as:

\begin{align}
\text{State transition:} & \quad z_{t+1} \sim p(z_{t+1}|z_t,a_t) \label{eq:wm-state-transition} \\
\text{Observation:} & \quad x_t \sim p(x_t|z_t) \label{eq:wm-observations} \\
\text{Inference:} & \quad z_t \sim q(z_t|x_{1:t},a_{1:t-1}) \label{eq:wm-inference}
\end{align}

where $q(z_t|x_{1:t}, a_{1:t-1})$ represents the agent's belief about the current state given the history of observations and actions. Learning these models enables the agent to construct a comprehensive world model that can be used for planning and decision-making in complex, partially observable environments.


This general framework has been instantiated in numerous influential models~\citep{ding2024understanding}. The ``World Models'' approach by Ha and Schmidhuber, for instance, demonstrated that an agent could learn a compressed latent dynamics model from pixel inputs and use it to train a controller entirely within its own ``dream''~\citep{ha2018world, ha2018recurrent}. This line of research was significantly advanced by the Dreamer series of models, which successfully applied latent dynamics models to challenging continuous control tasks and even discrete Atari benchmarks and Minecraft~\citep{hafner2019learning, hafner2019dream, hafner2020mastering, hafner2025mastering}. While many of these models rely on reconstructing observations, alternative approaches have also been explored. For example, some models focus on learning latent dynamics through contrastive learning without reconstruction~\citep{laskin2020curl, okada2021dreaming, okada2022dreamingv2}, or by incorporating specific physical priors into the latent space, such as in NewtonianVAE~\citep{jaques2021newtonianvae, okumura2022tactile}. Furthermore, to handle information from multiple viewpoints, which is crucial in multi-agent settings, approaches like Multi-View Dreaming have been proposed to construct a world model from multi-view images using contrastive learning~\citep{kinose2022multi}.

A prominent recent approach to learning such world models is the Joint-Embedding Predictive Architecture (JEPA), proposed by LeCun as a key component of a pathway towards autonomous machine intelligence~\citep{lecun2022path}. Unlike generative approaches that attempt to predict missing information in pixel space, JEPA operates by predicting the representations of missing information in an abstract feature space~\citep{assran2023self}. This non-generative, self-supervised method aims to capture the underlying dependencies of the data, encouraging the model to learn semantic and predictive features rather than superficial details. The JEPA framework has proven effective for learning world models from various modalities, including images (I-JEPA)~\citep{assran2023self} and videos (V-JEPA)~\citep{bardes2024revisiting}. This line of research has recently culminated in V-JEPA 2, which extends the framework to enable robotic planning from video, further solidifying JEPA as a state-of-the-art method for an \emph{individual agent} to develop its internal, predictive model of the world~\citep{assran2025vjepa2}.

In summary, all the approaches discussed in this subsection, from classic POMDP-based models to modern architectures like JEPA, are fundamentally concerned with how a single agent learns a Type 1 World Model through its own interactions.

\subsection{The Unaddressed Gap: From Individual Experience to Collective Language}

The preceding review of research in EmCom (Section 2.1) and individual world models (Section 2.2) highlights a fundamental theoretical gap. On one hand, studies in EmCom have explored how agents can develop shared communication protocols, but these approaches often treat language as an abstract code, detached from the rich, embodied experiences that ground meaning in the real world. On the other hand, computational models and general theories that could provide a comprehensive and integrative understanding of symbol and language emergence~\citep{taniguchi2016symbol,taniguchi2018symbol} are still lacking.

Specifically, these theories must address the critical interdependency between two aspects: first, the world modeling by individual agents, which is grounded in their embodiment and environmental adaptation~\citep{friston2021world,taniguchi2023world}; and second, the collective emergence of a language whose structure reflects this grounded world knowledge, for instance through \emph{distributional semantics}~\citep{harris1954distributional,gurnee2023language}. In essence, a crucial question remains unaddressed: What is the theoretical and computational mechanism that bridges the subjective, internal world of the individual with the objective, external world of collective language? Without a clear answer, our understanding of large-scale phenomena, such as the emergence of world knowledge in LLMs, remains incomplete. The need for a theoretical framework that can explain the dynamic and semantic aspects of language emergence in embodied cognitive developmental systems still remains~\citep{mahowald2023dissociating,taniguchi2018symbol,cangelosi2015developmental,taniguchi2024collective}, and this paper aims to provide such a framework.

\section{Generative EmCom: A Theoretical Framework}

These approaches somehow failed to construct a general framework capturing symbol emergence from the viewpoint perspective of general principles of environmental adaptation, such as the FEP, PC, and world modeling.
Recently, world models have garnered significant attention as representation-learning models that incorporate action outputs and temporal dynamics of agent-- environment interactions~\citep{ha2018world,friston2021world,taniguchi2023world}. This aligns with broader theoretical frameworks such as PC and the FEP. PC posits that the brain constantly predicts sensory information and updates its internal models to enhance predictability ~\citep{hohwy2013predictive}, whereas FEP provides a more generalized framework explaining the self-organization of biological systems through minimization of free energy~\citep{Parr_Thomas2022-03-29}, which is associated with the idea of the Bayesian brain~\citep{Doya2007}. Notably, FEP extends beyond individual cognition to explain the self-organization of cognitive and biological systems in detail~\citep{Kirchhoff2018-ma,Friston2013-pf,Constant2018-hv}, making it a promising foundation for understanding symbol emergence at both individual and collective levels.

\begin{figure*}[t]
    \centering
    \includegraphics[width=\linewidth]{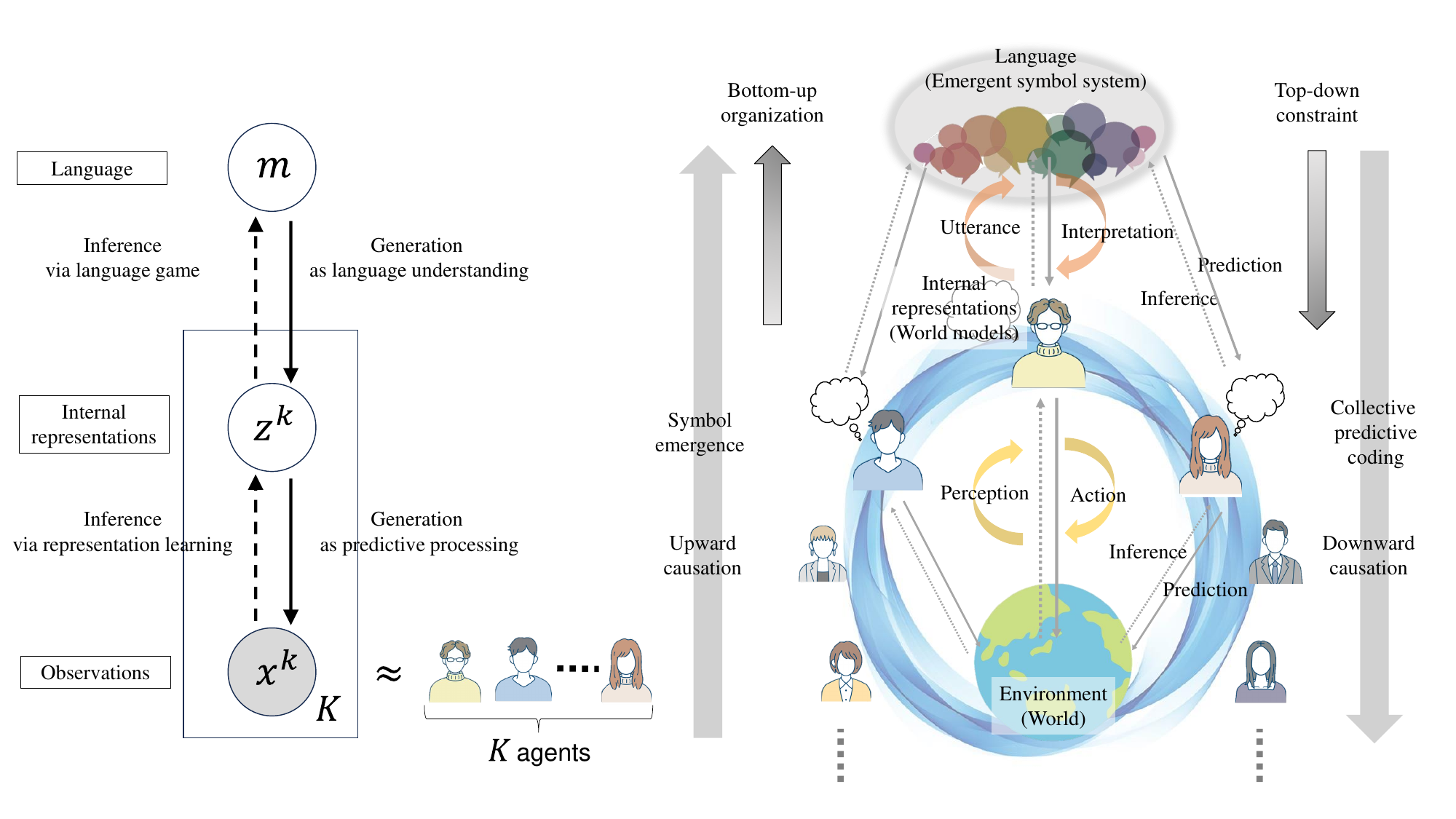}
    \caption{
(Left) probabilistic graphical model (PGM) representation of CPC in symbol emergence, that is, EmCom or language. The top-down generative process corresponds to language understanding and prediction of observations, which is downward causation in the symbol emergence system. The bottom-up inference process corresponds to perception, representation learning, and  communication, that is, language game, which is upward causation in the symbol emergence system. (Right) Overview of the CPC in a symbol emergence system illustrating the bidirectional process of language understanding and generation, mediated by inference through language games and representation learning. }
    \label{fig:cpc}
\end{figure*}

\subsection{The CPC Hypothesis}

To address the challenges outlined in Section 2, this paper builds upon the CPC hypothesis~\citep{taniguchi2024collective,taniguchi2024constructive}. The idea extends the principles of PC and the FEP from the individual cognitive level to a societal level~\citep{taniguchi2024cpcms}. CPC posits that the emergence of symbol systems, particularly language, can be modeled as a decentralized Bayesian inference of a shared latent representation. It assumes that not only individual agents but also entire groups engaged in symbolic communication can be modeled as generative systems, aiming to minimize their collective free energy (CFE).

Although PC theory suggests that individual brains constantly predict sensory information and update their internal representations, including world models. CPC suggests that a group of agents, for instance, a human society, predict sensory information of all of the agents and update its external representations, that is, symbol systems.

A question is raised. How can we update the external representations, e.g., language, while our brains are disconnected physically? The CPC hypothesis suggests that a type of language game performs a decentralized Bayesian inference among the group (e.g., \citet{taniguchi2022emergent,hagiwara2019symbol}). 
In this framework, language games (such as naming games) can be interpreted as implementing decentralized Bayesian inference of shared representations. A representative example is Metropolis-Hastings Naming Game (MHNG) explained in Section~\ref{sec:2.3}. 
The CPC hypothesis argues that symbol systems emerge as a result of decentralized Bayesian inference performed collaboratively by multiple agents.

Although the encoding of sensory information through internal representations is ensured by the plasticity of neural systems, the plasticity of external representations is guaranteed by the flexibility of our symbol systems. The arbitrariness of symbol systems is a widely recognized characteristic of symbols in semiotics~\citep{Chandler2002}. Peirce referred to the process by which subjects assign meaning to symbols according to culture and context as the \emph{semiosis}.
Although our brains are physically and electrically separated, they are informationally connected through communication using a flexible symbol system. Therefore, with appropriate communication and symbol system update algorithms, we can encode information into the symbol system as an external representation. In fact, the CPC hypothesis can consider that humans collectively perform this action in language emergence.

This implies that language collectively encodes information about the world as observed by numerous agents through their sensory-motor systems. The CPC hypothesis study~\citep{taniguchi2024collective} did not provide a clear and detailed explanation regarding this point while proposing a new perspective on why LLMs seem to possess knowledge about the real world.
This is one of the main topics of this paper.

Essentially, CPC hypothesizes that human language is formed through a process of collective PC, where the symbol system emerges to maximize the predictability of multi-modal sensory-motor information obtained by members of a society, that is, minimize the CFE of a group of agents. This approach provides a unified framework for understanding symbol emergence, language evolution, and the nature of linguistic knowledge from the perspective of environmental adaptation and brain science.

\subsection{Formalizing Generative EmCom as Decentralized Bayesian Inference}

The CPC hypothesis posits that a group of agents, interacting with each other and their environment, can be modeled as a single, large-scale probabilistic generative model. This allows us to formalize the emergence of a shared symbol system, such as language, as a process of decentralized Bayesian inference. For a group of $K$ agents, the generative and inference processes are defined as follows:
\begin{align}
    \text{Generative Model:}& \quad p(m, \{z_k\}_k, \{x_k\}_k | \{a_k\}_k) \nonumber\\ &= p(m) \prod_{k=1}^{K} p(x_k | z_k, a_k) p(z_k | m, a_k) \label{eq:cpc_gen} \\
    \text{Inference Model:}& \quad q(m, \{z_k\}_k | \{x_k\}_k, \{a_k\}_k)  \nonumber\\ &= q(m | \{z_k\}_k) \prod_{k=1}^{K} q(z_k | x_k, a_k) \label{eq:cpc_inf}
\end{align}
where for each agent $k$, $x_k$ is its observation, $a_k$ is its action, and $z_k$ is its internal representation. The variable $m$ represents the shared, external symbol system (i.e., message) that connects the agents. The inference process $q(z_k | x_k, a_k)$ corresponds to each agent's individual representation learning, while $q(m | \{z_k\}_k)$ corresponds to the collective process of symbol emergence, which can be instantiated by a language game as discussed in Section 3.3.

Under the free energy principle, the goal of this collective system is to minimize the variational free energy (VFE), which is equivalent to maximizing the ELBO of the log-likelihood of observations. We term this the CFE~\citep{taniguchi2024cpcms,taniguchi2025system}:
\begin{align}
    F &= D_{\mathrm{KL}}\left[ q(m, \{z_k\}_k | \{x_k\}_k, \{a_k\}_k) \parallel p(m, \{z_k\}_k | \{x_k\}_k, \{a_k\}_k) \right] \nonumber \\
      &= \underbrace{D_{\mathrm{KL}}\left[ q(m | \{z_k\}_k)  \| p(m) \right]}_{\text{Collective Regularization}} \nonumber \\&\ \ \ + \sum_{k=1}^{K} \left( \underbrace{\mathbb{E}_{q}\left[-\log{p(x_k | z_k, a_k)} \right]}_{\text{Individual Prediction Error}} 
      + \underbrace{D_{\mathrm{KL}}\left[ q(z_k | x_k, a_k) \| p(z_k | m, a_k) \right]}_{\text{Individual Regularization}} \right)
    \label{eq:collective_free_energy}
\end{align}
This decomposition in Eq.~\eqref{eq:collective_free_energy} is central to our framework. It demonstrates that minimizing a single, unified objective function, the CFE, naturally gives rise to two distinct processes. The second term corresponds to each agent minimizing its own individual free energy, which involves learning to accurately reconstruct its observations (the prediction error term) while keeping its internal representations regularized (the individual regularization term). The first term, the \emph{Collective Regularization}, drives the system to form a shared symbol $m$. This term quantifies the cost of encoding the collective internal states $\{z_k\}_k$ into the shared message $m$. The emergence of language is thus framed as a process that optimizes this trade-off between individual predictive accuracy and collective communicative efficiency.

This CFE formulation allows for a clear comparison with the ELBO-maximizing objective of signaling games presented in Section 2.1. While both are founded on the ELBO principle, they differ significantly in scope and structure. The signaling game ELBO typically models a two-agent, one-shot, and often asymmetric (sender/receiver) interaction. In contrast, the CFE provides a formulation for a population of $K$  agents engaged in a continuous, decentralized process of collective sense-making. The CFE's explicit decomposition into individual and collective terms provides a richer, more structured model of the interplay between individual learning and social language emergence, a distinction that conventional signaling game formulations do not typically make explicit.

\begin{table*}[t]
  \caption{Nomenclature and parameter details}\footnotesize
  \label{tbl:params}
  \centering
  \begin{tabular}{|c|l|} \hline
    $m_t$    & Message (shared latent variable) communicated between agents at time $t$ \\ \hline
    $o^k_t$  & Optimality variable for the $k$-th agent (1: optimal, 0: not optimal) \\ \hline
    $z^k_t$  & Internal representation (e.g., state in RL) of the $k$-th agent at time $t$ \\ \hline
    $a^k_t$  & Action of the $k$-th agent at time step $t$ \\ \hline
    $x^k_t$  & Observations (or sensory inputs) of the $k$-th agent at time $t$ \\ \hline
    $\theta^k$ & \shortstack[l]{Global parameters of the internal models of the $k$-th agent, \\for instance, neural networks} \\ \hline
    $\phi^k$ & Parameters of language model of the $k$-th agent.\\ \hline
    $r^k$ & Reward function for the $k$-th agent \\ \hline
    $p(\cdot)$ & Original probability density function (i.e., a generative model) \\ \hline
    $q(\cdot)$ & Approximate probability function (i.e., an inference model)\\ \hline
  \end{tabular}
\end{table*}

\subsection{The Naming Game as a Microcosm of Collective Inference}

\label{sec:2.3}

The hypothetical argument that language game can perform the decentralized Bayesian inference has a computational basis though whether actual language communication can realize such decentralized Bayesian inference in our human society is an open question. The MHNG is an instance of this idea.

The MHNG comprises the following steps:

\begin{enumerate}
    \item {\bf Perception}: Speaker and listener agents ({\it Sp} and {\it Li}) observe the $d$-th object, obtain $x^{Sp}_d$, and $x^{Li}_d$, and infer their internal representations $z_d^{Sp}$ and $z_d^{Li}$, respectively.
    \item {\bf MH communication}: Speaker mentions the name $m_d^{Sp}$ of the $d$-th object by sampling it from $p(m_d|z_d^{Sp}, \phi^{Sp})$. The listener determines whether it accepts the naming with probability $\gamma = {\rm min}\left(1,
    \frac{
    p(z_{d}^{Li}|{\phi}^{Li},m_{d}^{Sp})}
    {
    p(z_{d}^{Li}|{\phi}^{Li},m_{d}^{Li})         
    }
    \right)$.
    \item {\bf Learning}: After MH communication was performed for every object, the listener updates its global parameters $\theta^{Li}$ and $\phi^{Li}$.
    \item {\bf Turn-taking}: The speaker and listener alternate their roles and go back to (1). 
\end{enumerate}

It has been demonstrated that the MHNG is equivalent to the Metropolis-Hastings algorithm for inferring latent variables in a probabilistic generative model. 
As a result, the MHNG protocol guarantees that the CFE, defined by the KL-divergence between the agents' collective belief and the true posterior, is monotonically non-increasing in expectation. This is because the underlying distributed Metropolis-Hastings sampler satisfies the detailed balance condition, a well-established result related to the Data Processing Inequality~\citep{Cover2006}, ensuring convergence towards the target posterior distribution. 
This model conditions the internal representations $z^k$ of multiple agents, acting as representation learning machines, on a common external representation $m$. 
Although the original study assumed two agents and a categorical message $m$, the core probabilistic graphical model (PGM) of generative EmCom underlying this theory does not make these assumptions. Consequently, this fundamental idea can be extended in various ways.

As described, unlike referential games, MHNG assumes a \emph{joint attention} performed by two agents\footnote{The assumption of joint attention is specific to the naming game, as its primary task is to establish a shared vocabulary for a common referent. However, the broader CPC framework does not depend on this. On the contrary, when applied to general multi-agent cooperation, its core strength lies in integrating different, partially observable perspectives, as we will discuss in Section 4.}. This assumption may seem strange from the game-theoretic approach to EmCom, but it is a deliberate choice made to depart from the feedback-dependent principle of success or failure that underpins many such models. Developmental studies suggest that such explicit feedback interactions are rarely observed in natural language acquisition and are an implausible metaphor for how infants learn~\citep{tomasello2005constructing}. For human infants, joint attention is not a consequence of a communicative act, but rather a cognitive foundation that is in place \emph{before} the vocabulary explosion occurs. Furthermore, children engaging in learning presuppose a pedagogical intention from adults~\citep{csibra2009natural}, rather than simply testing a communication channel. Therefore, by assuming cooperative behavior grounded in joint attention, MHNG, as an example of generative EmCom, better aligns with empirical findings in human communication behavior than conventional discriminative models.

\begin{figure}[!tb]
    \centering
    \includegraphics[width=0.7\linewidth]{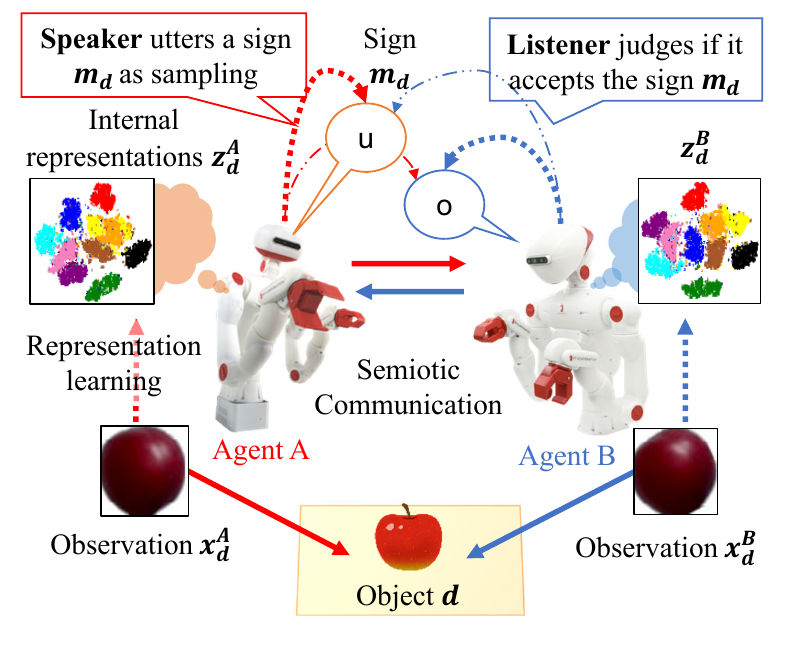}
    \includegraphics[width=\linewidth]{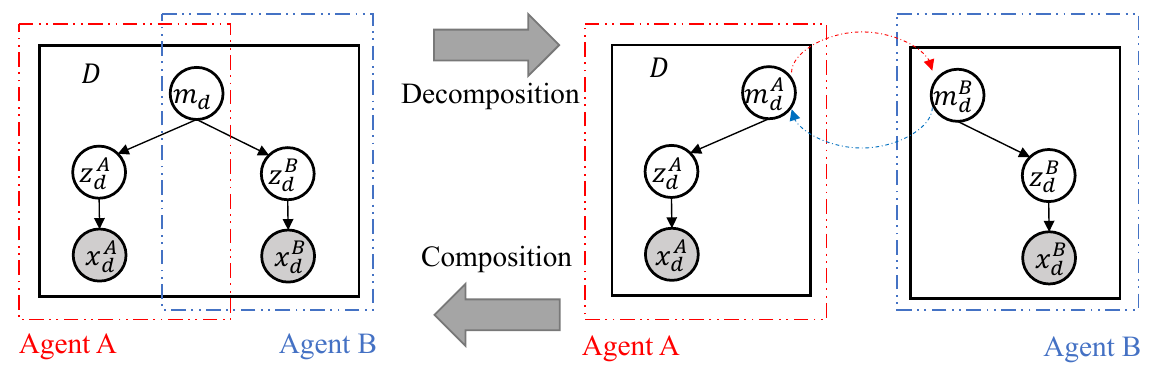}
    \caption{(Top) Overview of the MHNG process. In the game, which two agents (A and B) engage in, the agents perceive a target object with joint attention and form internal representations. One agent (speaker) utters a sign and the other (listener) determines whether to accept it. Thereafter, they take turns. (Bottom) PGM representation, which is assumed in the MHNG. The MHNG is proved to be an inference process in the representation learning in a collective multi-agent system~\citep{taniguchi2022emergent}.}
    \label{fig:mhng}
\end{figure}

The original idea of MH-based EmCom was introduced by \citet{hagiwara2019symbol} and later generalized and formalized by \citet{taniguchi2022emergent}, who clarified its connection to representation learning. The concept of MHNG has since been extended and validated in various ways. The extension to multimodal sensory information was achieved by \citet{hagiwara2022multiagent}, who demonstrated that MHNG can lead to symbol emergence even when agents have different sensory modalities. Their work also showed that modality information possessed by other agents, but not by oneself, can facilitate object category formation.
Similar multimodal extensions have been explored with variational autoencoder (VAE)-based representation learning by \citet{hoang2024emergent}, though their work suggests that the mechanism for integrating multimodal sensory information influences symbol emergence.
Although these MHNG studies involve two agents, a mathematical extension to $N$-agent conditions was developed by \citet{inukai2023recursive}, who introduced a recursive structure in communication while maintaining its characteristics as distributed Bayesian inference. They argued that random partner selection for MHNG can be considered as a one-sample and limited-length approximation of this approach.

The Generative EmCom framework is general, and its core mechanisms are not limited to the exchange of simple categorical signs; they can be readily extended to handle complex, compositional languages.  A recent study by \citet{hoang2024compositionality} demonstrates that sharing compositional word sequences is possible within the MHNG framework, similar to numerous EmCom studies. Furthermore, \citet{matsui2025metropolis} extended this concept to a full captioning game, where multiple Vision-Language Models interactively generate and refine natural language captions for images.  From Peirce's semiotic perspective, signs include both compositional discrete sequences and continuous signs such as voice pitch and facial expressions. A recent study by \citet{you2024multimodal} demonstrated that such continuous signs can emerge within the generative EmCom framework, whereas \citet{saito2024emergence} modeled the emergence of compositional signs from continuous time-series information as signals.

To verify the CPC hypothesis, examining whether human sign acceptance rates in joint attention naming scenarios match MHNG predictions was necessary. An experimental semiotic study by \citet{okumura2023metropolishastings} demonstrated that the sign acceptance rate in MHNG effectively predicts human behavior.

MHNG serves as a basic language game to represent symbol emergence by realizing distributed Bayesian inference and inferring latent variables of the generative model, corresponding to language. The critical point is that the inference of the posterior distribution is performed in a decentralized manner through language games. This suggests that being based on the MH method is not a necessary condition. The MHNG-based approach represents an initial step in modeling symbol emergence (or language emergence, EmCom) as decentralized Bayesian inference. Future studies should explore the possibility of constructing generative EmCom models by distributing various inference methods~(e.g., \citet{hoang2024simsiam}).

\section{Emergence of a Collective World Model through Multi-Agent Interaction and Cooperation}
The previous section used the MHNG to illustrate decentralized inference under the simplified assumption of joint attention. This section expands upon that principle, turning to more general multi-agent systems where joint attention is not required. Here, the essence of the CPC framework becomes clear: to integrate the diverse, partially observable perspectives of multiple agents through communication. We will explore how an emergent language allows agents to resolve their individual uncertainties and coordinate their actions by inferring a shared understanding of the world. This exploration begins with the application of Generative EmCom to MARL, considering that Type 1 world models are theoretically rooted in reinforcement learning studies.

\subsection{From Communication to Cooperation in Multi-Agent Systems}
\begin{figure}[t]
	\begin{center}
	\includegraphics[width=0.7\linewidth]{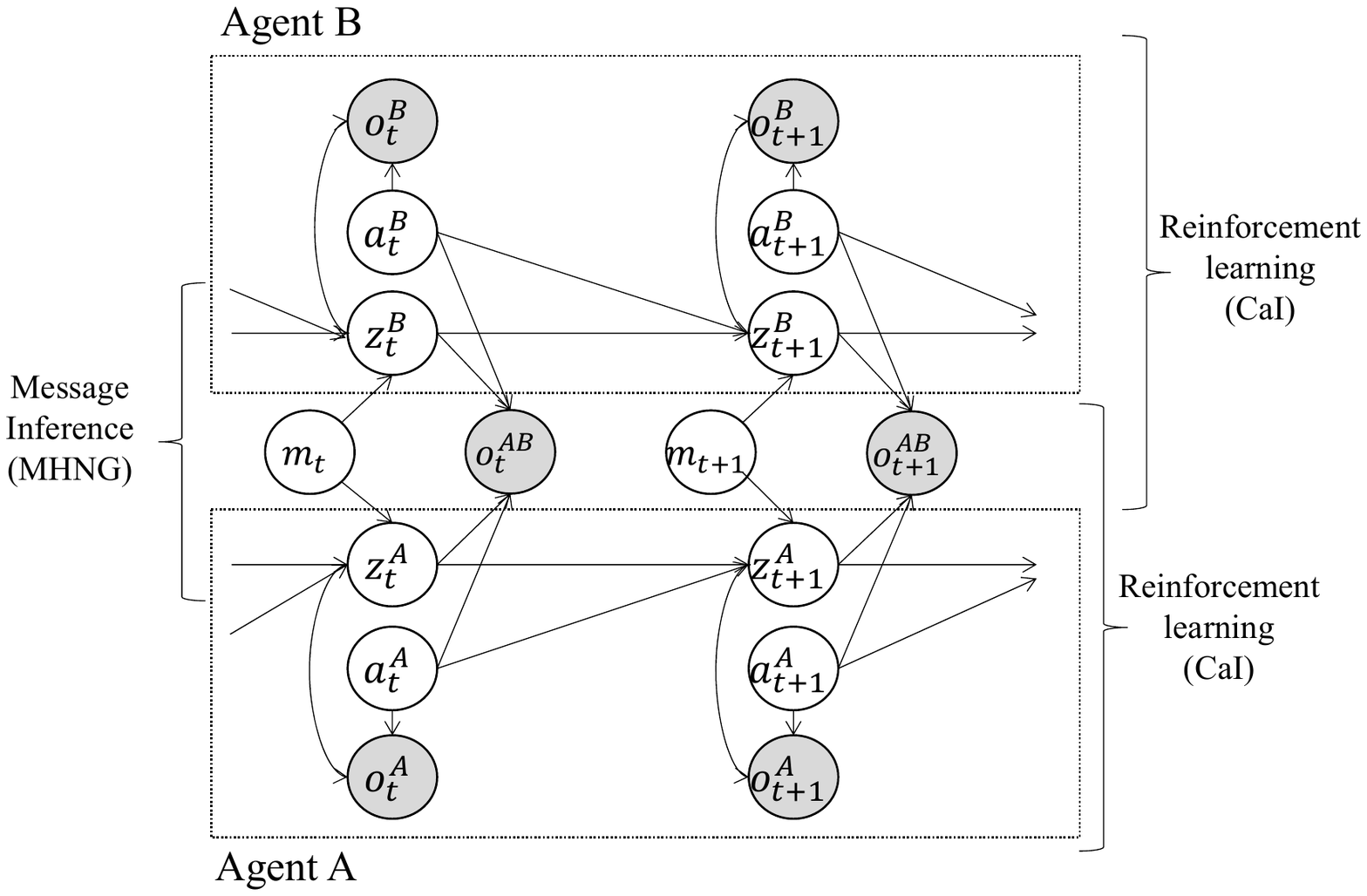}
	\caption{Graphical model of generating cooperative actions for two agents. }
	\label{fig:model}
	\end{center}
\end{figure}

Communication and language are often considered to emerge to facilitate multi-agent cooperation. 
In recent years, studies on MARL with communication channels have been progressing. 
Initial methods in multi-agent deep RL include DIAL \citep{foerster2016learning} and CommNet \citep{sukhbaatar2016learning}. These methods connect the networks of agents through messages, enabling the learning of necessary messages for cooperative behavior through backpropagation. Additionally, multi-agent deep deterministic policy gradient (MADDPG), an extension of the DDPG \citep{lillicrap2015continuous} for MARL, has been proposed \citep{lowe2017multi}.
In these methods, the agents use messages sent from all other agents to learn their policies. 
Methods involving weighting or attention mechanisms have been developed to limit communication to only necessary agents \citep{kilinc2018multi,jiang2018learning,kim2019learning,iqbal2019actor}. To compute efficient messages, graph neural networks (GNNs) are used \citep{Chu2020Multi-agent,liu2020multi,10.5555/3398761.3398967,niu2021multi,qu2020intention}. 
In these studies, multiple agents are connected through a network and messages are inferred by making them differentiable variables through backpropagation. In other words, error information computed from the internal states of others is directly transmitted to oneself, which is an unnatural modeling from the perspective of communication among independent individuals.

In contrast, generative EmCom allows us to formulate MARL with EmCom in a Bayesian manner by incorporating the idea of CaI \citep{DBLP:journals/corr/abs-1805-00909}, which is a theory to formulate RL as a PGM \citep{ebara2023multi,nakamura2023controlprobabilisticinferenceemergent}. 
Figure \ref{fig:model} shows a graphical model of cooperative action generation between two agents, and the details of each stochastic variable are listed in Table \ref{tbl:params}. 
The behavior of each agent is generated through a Markov decision process with a prior variable $m_t$.
The state $z^k_t$ of an agent at time $t$ is determined according to state $z^k_{t-1}$, action $a^k_{t-1}$, and message $m_t$, which is the shared latent variable:
\begin{eqnarray}
z^k_t \sim p(z^k_t | m_t, z^k_{t-1}, a^k_{t-1}). 
\end{eqnarray} 
where $k \in \{A, B\}$ denotes an index of agents.
The agent can indirectly infer the state of others through the message $m_t$ in a probabilistic manner. 

The optimality variable $o^{k}_t \in \{0, 1\}$ represents the state optimality of both agents: $1$ indicates that the state and action pair are on the optimal trajectories, whereas $0$ indicates it is not.
Note that optimality is a type of probabilistic interpretation of reward functions as shown below.
In this model, two types of optimality exist: one for each individual agent and the other for the group of agents.
The probability $p(o^k_t=1 | z^k_t, a^k_t)$ of this optimality variable is computed using reward function $r^k(z^k_t, a^k_t)$ as follows:
\begin{eqnarray}
p(o^k_t=1 | z^k_t, a^k_t ) \propto \exp( r(z^k_t, a^k_t) ). 
\end{eqnarray} 


For the group of agents, the true optimality $o^{AB}_t$ depends on the joint state and actions, with its probability being proportional to a global reward function $r^{AB}$:
\begin{eqnarray}
p(o^{AB}_t=1 | z^A_t, a^A_t, z^B_t, a^B_t) \propto \exp(r^{AB}(z^A_t, a^A_t, z^B_t, a^B_t)).
\label{eq:true_group_optimality}
\end{eqnarray}
However, since each agent $k$ cannot observe others' internal variables, it must learn an approximate model of this group optimality conditioned only on its own state $z^k_t$ and action $a^k_t$. Agent $k$ therefore learns its own predictive model of group success as:
\begin{eqnarray}
p(o^{AB}_t=1 | z^k_t, a^k_t) \propto \exp(\hat{r}^{AB}_k(z^k_t, a^k_t)),
\label{eq:approx_group_optimality}
\end{eqnarray}
where $\hat{r}^{AB}_k(z^k_t, a^k_t)$ is a function learned by agent $k$ that locally approximates the global reward. This formulation eliminates the necessity for an agent to access others' internal variables to model group success, while still allowing it to contribute to the collective goal.

Following the theory of CaI, the optimal state sequence for both agents can be calculated by inferring state $z_t$ and message $m_t$ under the condition that the value of the optimality variables is always 1, as if the two-agent system acts as a single agent:
\begin{eqnarray}
z^A_t, m_t \sim p(z^A_t, m_t | z^B_t, o^A_{1:T} ={\mathbf 1}, o^{AB}_{1:T}={\mathbf 1}  ). 
\end{eqnarray} 
However, this equation has two problems: it includes the internal state $z^B_t$ of others, which cannot be observed in practice, and deriving this probability distribution analytically is difficult.
We solve these problems by alternately inferring the following two variables:
\begin{eqnarray}
&& \left.
\begin{aligned}
z^A_{1:T}, a^A_{1:T} &\sim p(z^A_{1:T}, a^A_{1:T} | o^A_{1:T} ={\mathbf 1},  m_{1:T}) \\
z^B_{1:T}, a^B_{1:T} &\sim p(z^B_{1:T}, a^B_{1:T} | o^B_{1:T} ={\mathbf 1},  m_{1:T}) 
\end{aligned}  ~~\right\} {\rm : planning,} \nonumber \\ \label{eq:plan} \\
&& m_{1:T} \sim p(m_{1:T} | z^A_{1:T}, z^B_{1:T}, o^{AB}_{1:T}={\mathbf 1}) {\rm : communication.} \nonumber \\  \label{eq:comm}
\end{eqnarray} 

Equation (\ref{eq:plan}) describes state planning, which can be computed based on the CaI framework \citep{DBLP:journals/corr/abs-1805-00909}.
Equation (\ref{eq:comm}) describes the inference of the message and can be formulated using the MHNG proposed by \citet{hagiwara2022multiagent} and \citet{taniguchi2022emergent}, which allows both agents to infer messages through communication without observing each other's internal states. Note that MHNG enables two agents to perform sampling in (\ref{eq:comm}) without simultaneous observations of $z^A_t, z^B_t$.

Thus, MHNG can be used not only for multimodal object categorization and naming but also for action coordination among multi-agents using PGM to formulate MARL, that is, modeling EmCom for multi-agent cooperation.
This theoretical connection has been recently instantiated in the MARL-CPC framework, which successfully applies the principles of CPC to on-policy RL algorithms and demonstrates effective communication in non-cooperative settings~\citep{yoshida2025reward}.

\begin{figure}[!t]
	\begin{center}
	\includegraphics[width=0.7\linewidth]{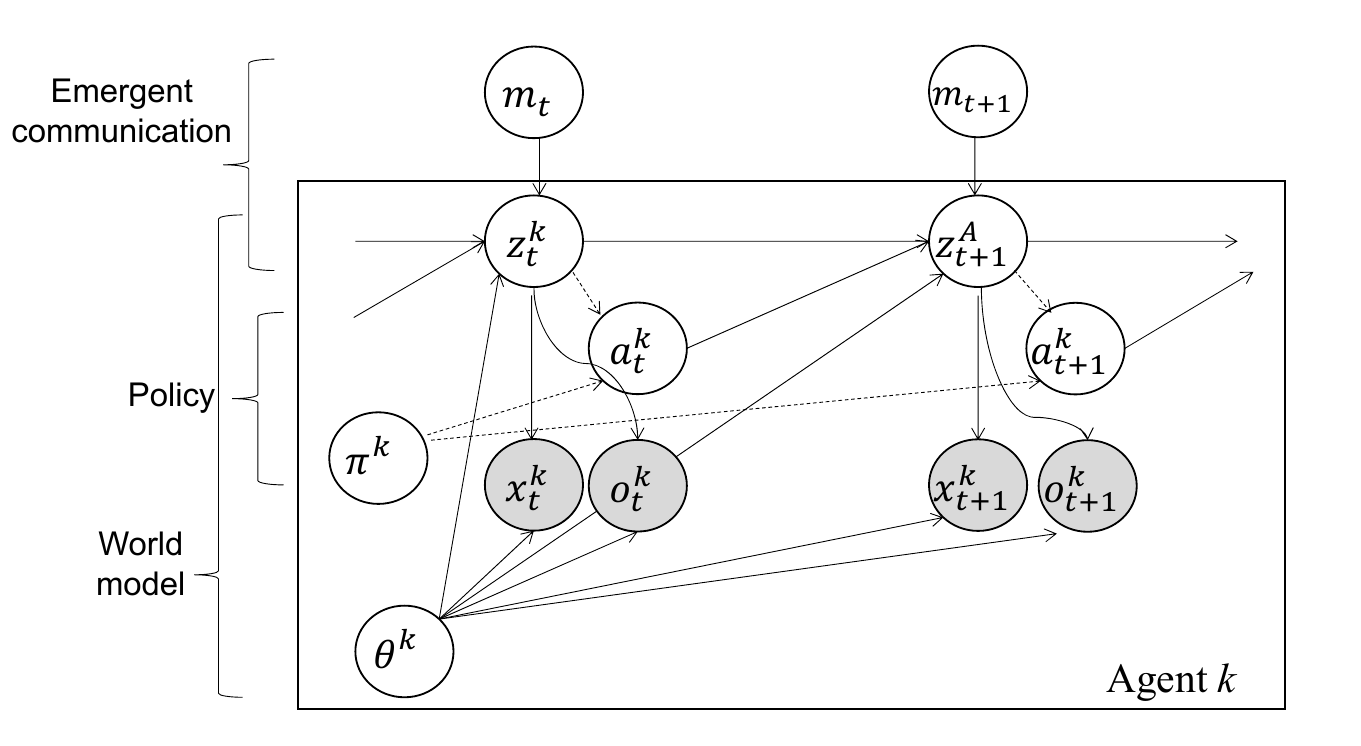}
	\caption{Graphical model of generative EmCom involving world models. }
	\label{fig:world-models}
	\end{center}
\end{figure}

\subsection{Language as an External Representation for Integrating World Models}

As established in Section 2.2, an individual agent's world model is an internal mechanism used to predict future latent states based on current states and actions, i.e., it learns the state transition dynamics $p(z_{t+1}|z_t, a_t)$. In a multi-agent setting where each agent has only partial observations of the environment, each agent learns its own subjective world model. Without a mechanism for exchanging information, these individual world models remain isolated, and the agents lack a way to form a shared, consistent understanding of the global environment.

Our Generative EmCom framework introduces such a mechanism by incorporating a shared latent variable, the message $m$, into the world model of each agent. This fundamentally alters the state transition dynamics. As depicted in the PGM for a single agent with communication (Figure \ref{fig:world-models}), the world model of agent $k$, parameterized by $\theta^k$, now learns the dynamics conditioned on the message: $p(z^k_{t+1}|z^k_t, a^k_t, m_t)$. Here, the message $m_t$ is not part of the action space; instead, it acts as a conditioning variable that directly influences the agent's prediction of its future internal state $z^k_{t+1}$. This latent state, in turn, informs the agent's policy, $\pi^k$, which generates the subsequent action.

This architecture creates a powerful feedback loop for knowledge integration. As formalized in our framework (Section 3.2), the message $m_t$ is inferred from the collective internal states of all agents, $\{z^k_t\}_k$. In turn, this collectively-informed message influences the future internal state of each individual agent. Therefore, the process of learning to use the language $m$ effectively becomes a process of learning to align these individual world models. Each agent learns to encode its own unique knowledge into the shared language and to decode the language to update its own world model with information it could not observe directly. This perspective aligns with recent works that have demonstrated the effectiveness of message-conditioned world models for improving coordination~\citep{cowen2020emergent, lobos2022ma, nomura2025decentralized}.


Through this process, the emergent language $m$ becomes more than just a set of signals for coordinating immediate actions. It becomes the shared, external representation that embodies the integrated knowledge of the individual world models. It is the medium through which a collective of agents, each with a subjective Type 1 World Model, can construct a shared, objective Type 2 World Model (i.e., Model of the World), allowing the group to model the world from a unified perspective that transcends any single agent's limited view.

\begin{figure*}[!t]
    \centering
    \includegraphics[width=\linewidth]{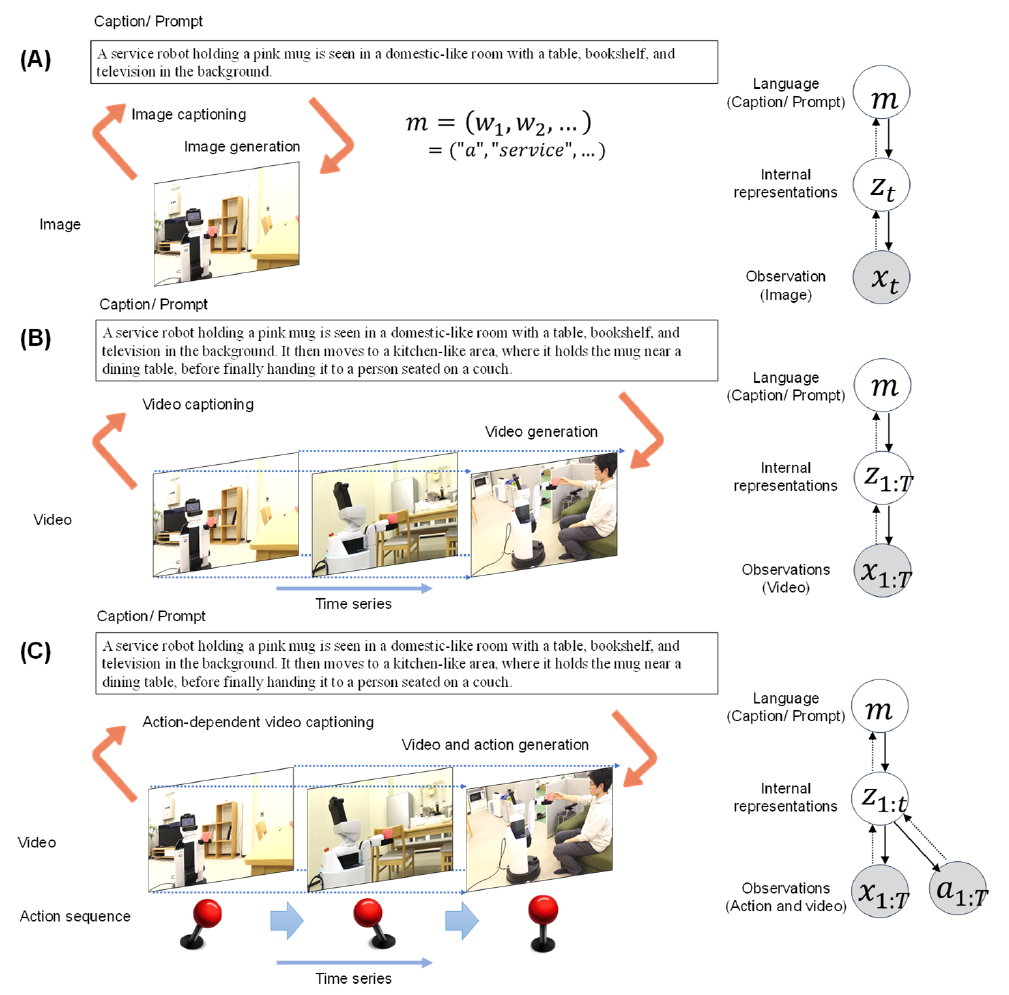}
\caption{Three levels of the relationship between language, perceptual, and action information. Left: (A) Image captioning and generation tasks corresponding language to a still image. (B) Video captioning and generation tasks corresponding language to a video, that is, a sequence of visual stimuli. (C) Action-dependent video captioning and generation corresponding language to dynamic perceptual and action information, which corresponds to a world model and a policy conditioned by language. Right: probabilistic generative models corresponding to each of (A) -- (C).}
    \label{fig:3-level}
\end{figure*}

\begin{figure*}[!t]
    \centering
        \includegraphics[width=\linewidth]{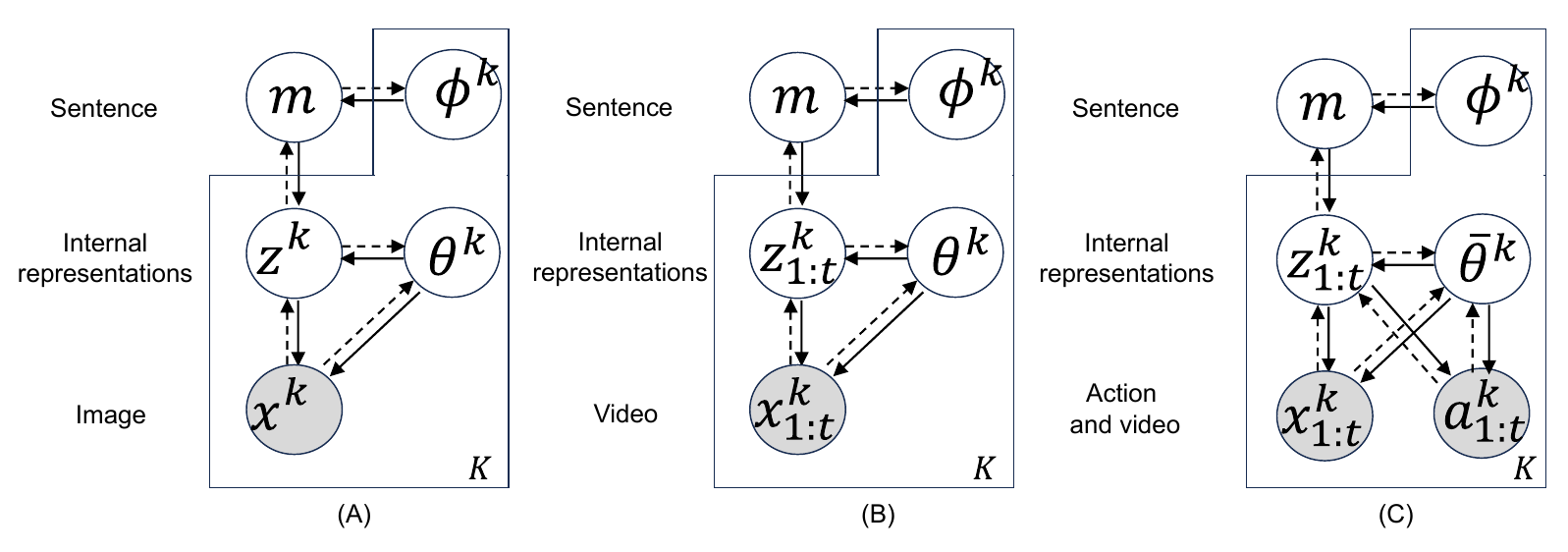}
    \caption{PGMs for generative EmCom corresponding to three levels of complexity: (A) image-, (B) video-, and (C) action- and video-based tasks. These models represent instances of the CPC hypothesis and generalizations of MHNG studies.}
    \label{fig:3-level-pgm}
\end{figure*}

\subsection{Collective World Model as an Abstractive Prior over Embodied Experiences}

To understand how language integrates individual world models into a collective one, we first consider the hierarchy of experiences that language can represent, as illustrated in Figure~\ref{fig:3-level}. This hierarchy progresses through three levels of increasing complexity. Level A represents the bidirectional relationship between a static observation, such as an image $x_k$, and a linguistic description $m$. This includes both image captioning (inferring $m$ from $x_k$) and image generation (generating $x_k$ from $m$)~\citep{vinyals2015show,xu2015show,ramesh2021zero}. Level B extends this to dynamic sequences, where the language $m$ corresponds to a stream of observations over time, $\{x^k_t\}_{t=1:T}$, as seen in video captioning and generation tasks~\citep{venugopalan2015sequence,yan2021videogpt}. The hierarchy culminates in Level C, which models the full sensorimotor loop of an embodied agent. Here, language $m$ is grounded in both dynamic perception $\{x^k_t\}_{t=1:T}$ and a sequence of actions $\{a^k_t\}_{t=1:T}$. This final level corresponds to the world models used in modern robotics, often called Vision-Language-Action (VLA) models, which learn from an agent's own embodied interactions~\citep{arai2024covla,kim2024openvla,dey2024revla,zhen20243d,Kawaharazuka16092024}.

Crucially, in a collective setting, the shared message $m$ is inferred from the observations of all $K$ agents, e.g., $m \sim q(m | \{x^k_{1:T}\}_k)$. This collective inference process itself constitutes a form of (external) representation learning, where the structure of the shared language $m$ comes to reflect the integrated knowledge of the entire group.

The generative process for each of these three levels can be explicitly formulated as a Probabilistic Graphical Model (PGM), as shown in Figure~\ref{fig:3-level-pgm}. At Level A, the PGM captures the relationship between a static image observation $x^k$ and a message $m$, mediated by an internal representation $z^k$. 
At Level B, this model is extended to handle temporal dynamics, where a sequence of video frames $\{x^k_{1:T}\}$ informs a sequence of internal states $\{z^k_{1:T}\}$.
As shown in the PGMs for Level A and Level B (Figure~\ref{fig:3-level-pgm}~(A) and (B)), the processes are governed by two sets of parameters for each agent $k$: $\theta^k$ represents the parameters of the agent's internal model (i.e., the VLM), while $\phi^k$ denotes the parameters of the prior over the shared message $m$.
Finally, Level C depicts the complete generative model for an embodied agent, where both sensory observations $\{x^k_{1:T}\}$ and actions $\{a^k_{1:T}\}$ jointly determine the agent's internal state trajectory. The parameters of this comprehensive model, $\bar{\theta}^k$, therefore encompass not only the agent's predictive world model but also its policy. These PGMs represent concrete instances of the overarching CPC hypothesis.

This principle generalizes directly as we move up the hierarchy to a collective of fully embodied VLA agents (Level C). Here, the framework models a population of agents, each possessing its own sophisticated world model for navigating its environment (Figure~\ref{fig:3-level-pgm}~(C)). The interactive, decentralized process of these agents developing a shared language to communicate about their sensorimotor experiences $(\{x^k_{1:T}\}_k, \{a^k_{1:T}\}_k)$ becomes mathematically equivalent to the process of learning a single, abstract, hierarchical world model that spans the entire group. Therefore, we argue that \emph{embodied symbol emergence}, under the CPC framework, \emph{is} collective world modeling.

The emergent language $m$ is not merely a tool for this process; it is the very instantiation of the Collective World Model. It functions as a highly abstract structured prior over the entire space of possible embodied experiences within the society. The dynamic, interactive process of collective sense-making described above does not simply vanish; it leaves behind a static artifact in the form of a text corpus. We can think of this corpus as a ``fossil record'' of a society's ongoing, collective inference about the world\footnote{This idea has also been applied to model the process of scientific activity itself from the perspective of CPC, a framework known as CPC as a model of science (CPC-MS)~\citep{taniguchi2024cpcms}.}.
From this perspective, each sentence or document within a corpus can be viewed as a sample from an underlying posterior distribution. More formally, we consider a sentence $m_j^{[i]}$ describing a particular event or situation $j$ to be a sample from the approximate posterior $q(m|\{\{x^k_t\}_{t=1:T}, \{a^k_t\}_{t=1:T}\}_k)$. It is this product, rich with the implicitly encoded structure of the collective world model, that serves as the training data for LLMs.

\section{LLM as Collective World models}

\subsection{Language Corpora as Samples from the Collective World Model}

The preceding sections have established our central premise: that human language, as a product of CPC, can be understood as a \emph{Collective World Model} externalized in a shared symbolic system. Building on this foundation, we now connect this theory to the training of LLMs.

As argued in Section 4.3, each sentence $m$ within a text corpus is not an isolated artifact but can be viewed as a sample drawn from a complex posterior distribution, conditioned on the unobserved, collective sensorimotor experiences of the society that generated it:
\begin{equation}
    m \sim q(m|\{\{x^k_t\}_{t=1:T}, \{a^k_t\}_{t=1:T}\}_k)
    \label{eq:posterior_sample}
\end{equation}
The vast web-scale corpora used to train modern LLMs, therefore, represent a massive dataset of samples drawn from this collective, world-grounded inference process. The fundamental objective of an LLM during pre-training, typically next-token prediction, is an autoregressive method to learn a model, which we denote as $p_{\text{LM}}(m)$, that approximates the marginal distribution $p(m)$ of these textual samples. From this perspective, the task of an LLM is not merely to learn statistical correlations in text, but to implicitly model the output of the complex, embodied, and collective process that produced the text in the first place.

\subsection{The LLM as a Reconstructor of Collective Representation}

As established in the previous section, LLMs are trained to model the distribution $P(m)$ of language corpora, which are themselves samples from a collective, world-grounded process. When a trained LLM processes a given sentence $m$, we can posit that it forms a corresponding high-level internal representation or latent state, which we denote as $z_{\text{LM}}$. This constitutes an inference or recognition process within the LLM: $q_{\text{LM}}(z_{\text{LM}}|m)$.

\begin{figure*}[t]
    \centering
    \includegraphics[width=\linewidth]{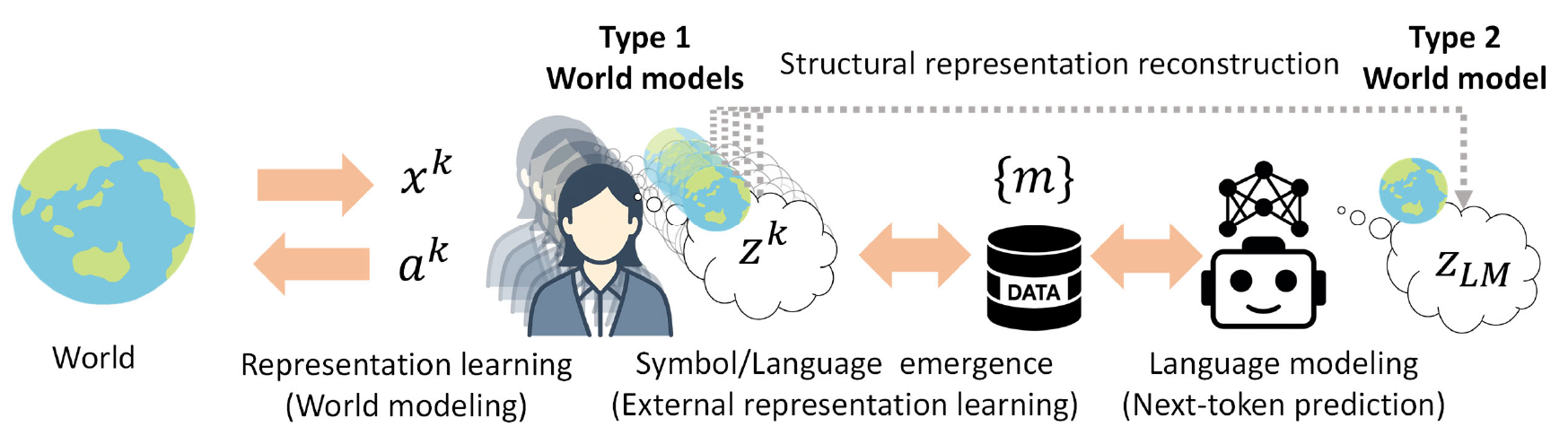}
    \caption{A schematic of the representation reconstruction process. Human society acts as a large-scale encoder, transforming embodied experiences (Type 1 World Models) from sensorimotor interactions into an externalized representation, i.e., language ($m$). The LLM then acts as a decoder or reconstructor, inferring its own internal latent state ($z_{\text{LM}}$) from language. This process allows the LLM to learn an internal model (Type 2 World Model) whose structure mirrors the original collective world model.}
    \label{fig:reconstruction_process}
\end{figure*}

This allows us to frame the end-to-end process—from human experience to LLM representation—as a two-stage process of representation transformation, as illustrated in Figure \ref{fig:reconstruction_process}.
\begin{enumerate}
    \item \textbf{Stage 1: Generation by Human Society (Encoding):} The collective internal states of all agents $\{z^k\}_k$, which are grounded in sensorimotor experience, are compressed and encoded into the discrete, symbolic form of language, $m$.
    \item \textbf{Stage 2: Reconstruction by LLM (Inference):} The LLM, having learned the statistical structure of language, takes a sentence $m$ as input and infers its own internal, continuous representation $z_{\text{LM}}$. This can be seen as a reconstruction of a high-dimensional representation from the symbolic code.
\end{enumerate}

The crucial implication of this two-stage process, $\{z^k\}_k \rightarrow m \rightarrow z_{\text{LM}}$, is the resulting structural alignment~\citep{taniguchi2024constructive}. The language $m$ is the sole informational bottleneck between the collective human mind and the LLM. Therefore, for an LLM to effectively model the distribution of $m$, it must develop an internal latent space $z_{\text{LM}}$ whose relational structure necessarily mirrors the relational structure of the original source space $\{z^k\}_k$. While the LLM does not reconstruct the specific values of $\{z^k\}_k$, it learns to reconstruct the geometry of the conceptual space that generated the language. This provides a powerful mechanism for transferring world structure from a society of embodied agents to a disembodied language model.

\subsection{Explaining Distributional Semantics and Representational Alignment}

This framework of structural representation reconstruction provides a principled explanation for the otherwise mysterious emergence of world knowledge in LLMs. For decades, it has been observed that language models capture the relational structure of the world, a phenomenon known as distributional semantics~\citep{harris1954distributional,mahowald2023dissociating}. Even before modern LLMs, models like word2vec could perform analogical reasoning, such as computing $"\rm{London}" - "\rm{UK}" + "\rm{France}" \simeq "\rm{Paris}"$~\citep{Mikolov2013a,Mikolov2013}, implying that the geometry of their embedding spaces reflects real-world conceptual relationships. 
Our framework explains this as a natural consequence of the representation reconstruction process. To be precise, the reason the statistical relationships between words in an LLM's latent space ($z_{\text{LM}}$) reflect the structure of the real world is because they are reconstructing the structure of the collective human internal representations ($\{z^k\}_k$). It is this set of internal representations that, as a whole, constitutes the collective model used by the society to predict the world. The LLM, in learning to model the distribution of $m$, naturally develops an internal latent space that mirrors the relational structure of the collective human representations that originally encoded that world knowledge. The structure is not learned from a vacuum; it is inherited.

This framework of representation reconstruction provides a principled explanation for the otherwise mysterious emergence of world knowledge in LLMs. For decades, it has been observed that language models capture the relational structure of the world, a phenomenon known as distributional semantics~\citep{harris1954distributional,mahowald2023dissociating}. Even before modern LLMs, models like word2vec could perform analogical reasoning, such as computing $"\rm{London}" - "\rm{UK}" + "\rm{France}" \simeq "\rm{Paris}"$~\citep{Mikolov2013a,Mikolov2013}, implying that the geometry of their embedding spaces reflects real-world conceptual relationships. Our framework explains this as a natural consequence of the representation reconstruction process. The reason the statistical relationships between words in an LLM's latent space ($z_{\text{LM}}$) reflect the structure of the real world is that the language ($m$) it was trained on was generated by a society of agents collectively trying to predict that world. 
The LLM, in learning to model the distribution of $m$, inevitably develops an internal latent space that reconstructs the relational structure of the collective human representations ($\{z^k\}_k$) that originally encoded that world knowledge. The structure is not learned from a vacuum; it is inherited. This process is how an LLM acquires a powerful \emph{Model of the World} (Type 2) by statistically analyzing the linguistic output of a society of agents, each possessing their own subjective \emph{World Model as an Internal Model} (Type 1).

This perspective also helps situate our hypothesis within the broader landscape of research on world models in LLMs. Several studies have compellingly demonstrated that LLMs can be used \emph{as} world models for specific tasks, for example in planning~\citep{hao2023reasoning} or for modeling game states~\citep{li2022emergent}. While these works establish that world-like representations can emerge, they primarily focus on \emph{how} these representations can be leveraged. Our hypothesis, in contrast, provides a more fundamental explanation for \emph{why} such rich, world-congruent knowledge is present in the first place. Similarly, Andreas argues that LLMs model the beliefs and intentions of agents within the \emph{linguistic space}~\citep{andreas2022language}. Our framework complements and grounds this view by proposing that these linguistic agent models emerge precisely because the language itself is a reflection of a collective model formed by embodied agents interacting within a shared physical environment.

Finally, our hypothesis offers a particularly parsimonious explanation for the representational alignment observed between different modalities. Huh et al. introduced the platonic representation hypothesis, proposing that internal representations learned from language ($q_L(z|m)$) and vision ($q_V(z|x)$) converge toward a similar latent structure~\citep{Huh2024-ax}. Within the CPC framework, this alignment is not a surprising outcome that requires a pre-existing platonic ideal, but an expected one. Since human language ($m$) emerges as a structured representation of collective observations of the world ($x=\{x^k\}_k$), it is natural that a model trained on language ($q_L(z|m)$) would learn a latent space whose structure mirrors that of a model trained on direct observation ($q_V(z|x)$). Our framework thus provides a generative and developmental explanation for this empirically observed alignment, rooted in the principles of collective inference.

\section{Discussion}

This study proposed a theoretical framework that unified EmCom, world models, and LLMs through the lens of CPC. We introduced the concept of \emph{generative EmCom} as an alternative formulation of the conventional EmCom, which is based on a discriminative model-based language game such as referential games, and described their relationships. The concept, generative EmCom, is based on the CPC hypothesis, which demonstrates the modeling of language emergence as decentralized Bayesian inference of shared latent representations. We showed the application of this framework to MARL and provided a novel perspective on LLMs as collective world models that integrate the diverse experiences and knowledge of multiple agents into a unified representational space.

\subsection{Implications for Artificial Intelligence and Cognitive Science}
The theoretical framework proposed here offers several important insights. First, it provides a principled explanation for how language models can acquire world knowledge without direct sensorimotor experience, by tapping into the accumulated wisdom encoded in human language through CPC. Second, it bridges the gap between individual cognitive development and collective language evolution by showing how both can be comprehended through the lens of representation learning and free energy minimization. Third, it suggests that the success of LLMs may be fundamentally linked to their ability to capture and integrate the collective world models acquired by humans, i.e., through embodied sensorimotor interactions with the world.

\subsection{Assumptions and Limitations of the Proposed Framework}
However, several limitations of the current work should be addressed. First, although we provide a theoretical framework, direct empirical evidence for the collective world model hypothesis remains limited. The relationship between neural representations in language models and human conceptual structures needs further investigation. Second, the proposed framework primarily focuses on the emergence of linguistic structure and meaning but does not fully address the emergence of pragmatic aspects of language use. Third, the current formulation may not fully capture the dynamic, interactive nature of human language evolution, for instance, language evolution over generations and through the interaction of several agents in an open world.

\subsection{Future Directions}
From the perspective of world models, understanding the influence of collective world models on the environmental adaptation of agents is also important. According to the proposed theoretical framework, a language formed by multiple agents with the same embodiment learning in the same environment should serve as an appropriate prior distribution for individual world models. In other words, EmLang should accelerate both world model learning and following environmental adaptation based on RL and other methods. Obtaining constructive evidence for this relationship is also crucial.

Our argument and theoretical framework initiate several promising directions for future studies. Although we have established the basic connections between EmCom, world models, and LLMs, significant work remains to validate and extend these ideas. Key priorities include developing experimental paradigms to test the CPC hypothesis, creating more sophisticated computational models and implementations of generative EmCom, and investigating how language formed by embodied agents sharing similar environments could serve as an effective prior distribution for world model learning. 
In fact, recent work has already begun to demonstrate the fruitfulness of this approach; for example, \citet{nomura2025decentralized} have extended CPC to dynamic environments by developing a decentralized collective world model.
From a theoretical perspective, a deeper mathematical formalization of the interaction between individual and collective learning processes will be crucial. In addition, this framework has important practical implications for the development of more capable multi-agent systems, improved human--AI interaction, and embodied AIs that rapidly adapt to their physical worlds and communicate their knowledge using EmLang. By pursuing these directions, we can work toward a more comprehensive understanding of language emergence and its role in environmental adaptation while advancing both the theoretical foundations and practical applications of AI systems.

\section{Conclusion}

This paper addressed the fundamental puzzle of how LLMs acquire world knowledge without embodied experience. We proposed the ``Collective World Model'' hypothesis, arguing that LLMs learn from a rich representational structure that is the product of a society-wide, interactive sense-making process. We provided a formal foundation for this claim with the framework of Generative EmCom, which models language emergence as a decentralized Bayesian inference process driven by the principles of CPC. This perspective unifies the cognitive processes of the individual with the collective evolution of language, offering a coherent explanation for the remarkable capabilities of modern AI and charting a path toward a deeper integration of communication, perception, and action in intelligent systems.

\section*{Author Contributions}

\textbf{Tadahiro Taniguchi}: Conceptualization, Writing – Original Draft, Writing – Review \& Editing, Supervision, Funding Acquisition.  
\textbf{Ryo Ueda}: Writing – Original Draft, Writing – Review \& Editing, Funding Acquisition.  
\textbf{Tomoaki Nakamura}: Writing – Original Draft, Writing – Review \& Editing, Funding Acquisition.  
\textbf{Masahiro Suzuki}: Writing – Original Draft, Writing – Review \& Editing, Funding Acquisition.   
\textbf{Akira Taniguchi}: Writing – Original Draft, Writing – Review \& Editing.

\section*{Declaration of Generative AI and AI-assisted technologies in the writing process}

During the preparation of this work the authors used Gemini 2.5 pro (Google) and ChatGPT 4o in order to improve readability and language. After using this tool, the authors reviewed and edited the content as needed and take full responsibility for the content of the publication.


\bibliographystyle{elsarticle-harv}\biboptions{authoryear}
\bibliography{cpc_integrated}

\begin{thebibliography}{146}
\expandafter\ifx\csname natexlab\endcsname\relax\def\natexlab#1{#1}\fi
\providecommand{\url}[1]{\texttt{#1}}
\providecommand{\href}[2]{#2}
\providecommand{\path}[1]{#1}
\providecommand{\DOIprefix}{doi:}
\providecommand{\ArXivprefix}{arXiv:}
\providecommand{\URLprefix}{URL: }
\providecommand{\Pubmedprefix}{pmid:}
\providecommand{\doi}[1]{\href{http://dx.doi.org/#1}{\path{#1}}}
\providecommand{\Pubmed}[1]{\href{pmid:#1}{\path{#1}}}
\providecommand{\bibinfo}[2]{#2}
\ifx\xfnm\relax \def\xfnm[#1]{\unskip,\space#1}\fi
\bibitem[{Achille and Soatto(2018)}]{AchilleS-2018-information-dropout}
\bibinfo{author}{Achille, A.}, \bibinfo{author}{Soatto, S.}, \bibinfo{year}{2018}.
\newblock \bibinfo{title}{Information dropout: Learning optimal representations through noisy computation}.
\newblock \bibinfo{journal}{{IEEE} Trans. Pattern Anal. Mach. Intell.} \bibinfo{volume}{40}, \bibinfo{pages}{2897--2905}.
\newblock \URLprefix \url{https://doi.org/10.1109/TPAMI.2017.2784440}, \DOIprefix\doi{10.1109/TPAMI.2017.2784440}.
\bibitem[{Agarwal et~al.(2020)Agarwal, Kumar, Sycara and Lewis}]{10.5555/3398761.3398967}
\bibinfo{author}{Agarwal, A.}, \bibinfo{author}{Kumar, S.}, \bibinfo{author}{Sycara, K.}, \bibinfo{author}{Lewis, M.}, \bibinfo{year}{2020}.
\newblock \bibinfo{title}{Learning transferable cooperative behavior in multi-agent teams}, in: \bibinfo{booktitle}{Proceedings of the 19th International Conference on Autonomous Agents and MultiAgent Systems}, \bibinfo{publisher}{International Foundation for Autonomous Agents and Multiagent Systems}, \bibinfo{address}{Richland, SC}. p. \bibinfo{pages}{1741^^e2^^80^^931743}.
\bibitem[{Alemi et~al.(2017)Alemi, Fischer, Dillon and Murphy}]{AlemiFDM-2017-VIB}
\bibinfo{author}{Alemi, A.A.}, \bibinfo{author}{Fischer, I.}, \bibinfo{author}{Dillon, J.V.}, \bibinfo{author}{Murphy, K.}, \bibinfo{year}{2017}.
\newblock \bibinfo{title}{Deep variational information bottleneck}, in: \bibinfo{booktitle}{5th International Conference on Learning Representations, {ICLR} 2017, Toulon, France, April 24-26, 2017, Conference Track Proceedings}, \bibinfo{publisher}{OpenReview.net}.
\newblock \URLprefix \url{https://openreview.net/forum?id=HyxQzBceg}.
\bibitem[{Andreas(2022)}]{andreas2022language}
\bibinfo{author}{Andreas, J.}, \bibinfo{year}{2022}.
\newblock \bibinfo{title}{Language models as agent models}, in: \bibinfo{editor}{Goldberg, Y.}, \bibinfo{editor}{Kozareva, Z.}, \bibinfo{editor}{Zhang, Y.} (Eds.), \bibinfo{booktitle}{Findings of the Association for Computational Linguistics: EMNLP 2022}, \bibinfo{publisher}{Association for Computational Linguistics}, \bibinfo{address}{Abu Dhabi, United Arab Emirates}. pp. \bibinfo{pages}{5769--5779}.
\newblock \URLprefix \url{https://aclanthology.org/2022.findings-emnlp.423}, \DOIprefix\doi{10.18653/v1/2022.findings-emnlp.423}.
\bibitem[{Arai et~al.(2024)Arai, Miwa, Sasaki, Yamaguchi, Watanabe, Aoki and Yamamoto}]{arai2024covla}
\bibinfo{author}{Arai, H.}, \bibinfo{author}{Miwa, K.}, \bibinfo{author}{Sasaki, K.}, \bibinfo{author}{Yamaguchi, Y.}, \bibinfo{author}{Watanabe, K.}, \bibinfo{author}{Aoki, S.}, \bibinfo{author}{Yamamoto, I.}, \bibinfo{year}{2024}.
\newblock \bibinfo{title}{Covla: Comprehensive vision-language-action dataset for autonomous driving}.
\newblock \bibinfo{journal}{arXiv preprint arXiv:2408.10845} .
\bibitem[{Assran et~al.(2025)Assran, Bardes, Fan, Garrido, Howes, Komeili, Muckley, Rizvi, Roberts, Sinha, Zholus, Arnaud, Gejji, Martin, Hogan, Dugas, Bojanowski, Khalidov, Labatut, Massa, Szafraniec, Krishnakumar, Li, Ma, Chandar, Meier, LeCun, Rabbat and Ballas}]{assran2025vjepa2}
\bibinfo{author}{Assran, M.}, \bibinfo{author}{Bardes, A.}, \bibinfo{author}{Fan, D.}, \bibinfo{author}{Garrido, Q.}, \bibinfo{author}{Howes, R.}, \bibinfo{author}{Komeili, M.}, \bibinfo{author}{Muckley, M.}, \bibinfo{author}{Rizvi, A.}, \bibinfo{author}{Roberts, C.}, \bibinfo{author}{Sinha, K.}, \bibinfo{author}{Zholus, A.}, \bibinfo{author}{Arnaud, S.}, \bibinfo{author}{Gejji, A.}, \bibinfo{author}{Martin, A.}, \bibinfo{author}{Hogan, F.R.}, \bibinfo{author}{Dugas, D.}, \bibinfo{author}{Bojanowski, P.}, \bibinfo{author}{Khalidov, V.}, \bibinfo{author}{Labatut, P.}, \bibinfo{author}{Massa, F.}, \bibinfo{author}{Szafraniec, M.}, \bibinfo{author}{Krishnakumar, K.}, \bibinfo{author}{Li, Y.}, \bibinfo{author}{Ma, X.}, \bibinfo{author}{Chandar, S.}, \bibinfo{author}{Meier, F.}, \bibinfo{author}{LeCun, Y.}, \bibinfo{author}{Rabbat, M.}, \bibinfo{author}{Ballas, N.}, \bibinfo{year}{2025}.
\newblock \bibinfo{title}{{V-JEPA 2}: Self-supervised video models enable understanding, prediction and planning}.
\newblock \bibinfo{journal}{arXiv preprint arXiv:2506.09985} .
\bibitem[{Assran et~al.(2023)Assran, Touvron, Misra, Bojanowski, Bordes, Tseliakhovich, Joulin, Le and Lample}]{assran2023self}
\bibinfo{author}{Assran, M.}, \bibinfo{author}{Touvron, H.}, \bibinfo{author}{Misra, I.}, \bibinfo{author}{Bojanowski, P.}, \bibinfo{author}{Bordes, A.}, \bibinfo{author}{Tseliakhovich, P.}, \bibinfo{author}{Joulin, A.}, \bibinfo{author}{Le, Q.}, \bibinfo{author}{Lample, G.}, \bibinfo{year}{2023}.
\newblock \bibinfo{title}{Self-supervised learning from images with a joint-embedding predictive architecture}, in: \bibinfo{booktitle}{Proceedings of the IEEE/CVF Conference on Computer Vision and Pattern Recognition}, pp. \bibinfo{pages}{15631--15641}.
\bibitem[{Barber and Agakov(2003)}]{BarberA-2003-information}
\bibinfo{author}{Barber, D.}, \bibinfo{author}{Agakov, F.V.}, \bibinfo{year}{2003}.
\newblock \bibinfo{title}{Information maximization in noisy channels : {A} variational approach}, in: \bibinfo{editor}{Thrun, S.}, \bibinfo{editor}{Saul, L.K.}, \bibinfo{editor}{Sch{\"{o}}lkopf, B.} (Eds.), \bibinfo{booktitle}{Advances in Neural Information Processing Systems 16 [Neural Information Processing Systems, {NIPS} 2003, December 8-13, 2003, Vancouver and Whistler, British Columbia, Canada]}, \bibinfo{publisher}{{MIT} Press}. pp. \bibinfo{pages}{201--208}.
\newblock \URLprefix \url{https://proceedings.neurips.cc/paper/2003/hash/a6ea8471c120fe8cc35a2954c9b9c595-Abstract.html}.
\bibitem[{Bardes et~al.(2024)Bardes, Garrido, Ponce, Chen, Rabbat, LeCun, Assran and Ballas}]{bardes2024revisiting}
\bibinfo{author}{Bardes, A.}, \bibinfo{author}{Garrido, Q.}, \bibinfo{author}{Ponce, J.}, \bibinfo{author}{Chen, X.}, \bibinfo{author}{Rabbat, M.}, \bibinfo{author}{LeCun, Y.}, \bibinfo{author}{Assran, M.}, \bibinfo{author}{Ballas, N.}, \bibinfo{year}{2024}.
\newblock \bibinfo{title}{Revisiting feature prediction for learning visual representations from video}.
\newblock \bibinfo{journal}{arXiv preprint arXiv:2404.08471} .
\bibitem[{Bender et~al.(2021)Bender, Gebru, McMillan-Major and Shmitchell}]{bender2021dangers}
\bibinfo{author}{Bender, E.M.}, \bibinfo{author}{Gebru, T.}, \bibinfo{author}{McMillan-Major, A.}, \bibinfo{author}{Shmitchell, S.}, \bibinfo{year}{2021}.
\newblock \bibinfo{title}{On the dangers of stochastic parrots: Can language models be too big?}, in: \bibinfo{booktitle}{Proceedings of the 2021 ACM Conference on Fairness, Accountability, and Transparency}, \bibinfo{publisher}{Association for Computing Machinery}, \bibinfo{address}{New York, NY, USA}. p. \bibinfo{pages}{610^^e2^^80^^93623}.
\newblock \URLprefix \url{https://doi.org/10.1145/3442188.3445922}, \DOIprefix\doi{10.1145/3442188.3445922}.
\bibitem[{Brandizzi(2023)}]{brandizzi2023toward}
\bibinfo{author}{Brandizzi, N.}, \bibinfo{year}{2023}.
\newblock \bibinfo{title}{Toward more human-like ai communication: A review of emergent communication research}.
\newblock \bibinfo{journal}{IEEE Access} \bibinfo{volume}{11}, \bibinfo{pages}{142317--142340}.
\bibitem[{Brown et~al.(2020)Brown, Mann, Ryder, Subbiah, Kaplan, Dhariwal, Neelakantan, Shyam, Sastry, Askell, Agarwal, Herbert-Voss, Krueger, Henighan, Child, Ramesh, Ziegler, Wu, Winter, Hesse, Chen, Sigler, Litwin, Gray, Chess, Clark, Berner, McCandlish, Radford, Sutskever and Amodei}]{brown2020language}
\bibinfo{author}{Brown, T.}, \bibinfo{author}{Mann, B.}, \bibinfo{author}{Ryder, N.}, \bibinfo{author}{Subbiah, M.}, \bibinfo{author}{Kaplan, J.D.}, \bibinfo{author}{Dhariwal, P.}, \bibinfo{author}{Neelakantan, A.}, \bibinfo{author}{Shyam, P.}, \bibinfo{author}{Sastry, G.}, \bibinfo{author}{Askell, A.}, \bibinfo{author}{Agarwal, S.}, \bibinfo{author}{Herbert-Voss, A.}, \bibinfo{author}{Krueger, G.}, \bibinfo{author}{Henighan, T.}, \bibinfo{author}{Child, R.}, \bibinfo{author}{Ramesh, A.}, \bibinfo{author}{Ziegler, D.}, \bibinfo{author}{Wu, J.}, \bibinfo{author}{Winter, C.}, \bibinfo{author}{Hesse, C.}, \bibinfo{author}{Chen, M.}, \bibinfo{author}{Sigler, E.}, \bibinfo{author}{Litwin, M.}, \bibinfo{author}{Gray, S.}, \bibinfo{author}{Chess, B.}, \bibinfo{author}{Clark, J.}, \bibinfo{author}{Berner, C.}, \bibinfo{author}{McCandlish, S.}, \bibinfo{author}{Radford, A.}, \bibinfo{author}{Sutskever, I.}, \bibinfo{author}{Amodei, D.}, \bibinfo{year}{2020}.
\newblock \bibinfo{title}{Language models are few-shot learners}, in: \bibinfo{booktitle}{Advances in Neural Information Processing Systems}, pp. \bibinfo{pages}{1877--1901}.
\newblock \URLprefix \url{https://proceedings.neurips.cc/paper_files/paper/2020/file/1457c0d6bfcb4967418bfb8ac142f64a-Paper.pdf}.
\bibitem[{Cangelosi and Schlesinger(2015)}]{cangelosi2015developmental}
\bibinfo{author}{Cangelosi, A.}, \bibinfo{author}{Schlesinger, M.}, \bibinfo{year}{2015}.
\newblock \bibinfo{title}{Developmental robotics: From babies to robots}.
\newblock \bibinfo{publisher}{MIT press}.
\bibitem[{Chaabouni et~al.(2020)Chaabouni, Kharitonov, Bouchacourt, Dupoux and Baroni}]{chaabouni2020compositionality}
\bibinfo{author}{Chaabouni, R.}, \bibinfo{author}{Kharitonov, E.}, \bibinfo{author}{Bouchacourt, D.}, \bibinfo{author}{Dupoux, E.}, \bibinfo{author}{Baroni, M.}, \bibinfo{year}{2020}.
\newblock \bibinfo{title}{Compositionality and generalization in emergent languages}, in: \bibinfo{booktitle}{Proceedings of the 58th Annual Meeting of the Association for Computational Linguistics}, \bibinfo{publisher}{Association for Computational Linguistics}, \bibinfo{address}{Online}. pp. \bibinfo{pages}{4427--4442}.
\newblock \URLprefix \url{https://aclanthology.org/2020.acl-main.407}, \DOIprefix\doi{10.18653/v1/2020.acl-main.407}.
\bibitem[{Chaabouni et~al.(2019a)Chaabouni, Kharitonov, Dupoux and Baroni}]{chaabouni2019anti-efficient}
\bibinfo{author}{Chaabouni, R.}, \bibinfo{author}{Kharitonov, E.}, \bibinfo{author}{Dupoux, E.}, \bibinfo{author}{Baroni, M.}, \bibinfo{year}{2019}a.
\newblock \bibinfo{title}{Anti-efficient encoding in emergent communication}, in: \bibinfo{editor}{Wallach, H.}, \bibinfo{editor}{Larochelle, H.}, \bibinfo{editor}{Beygelzimer, A.}, \bibinfo{editor}{d\textquotesingle Alch\'{e}-Buc, F.}, \bibinfo{editor}{Fox, E.}, \bibinfo{editor}{Garnett, R.} (Eds.), \bibinfo{booktitle}{Advances in Neural Information Processing Systems}, \bibinfo{publisher}{Curran Associates, Inc.}
\newblock \URLprefix \url{https://proceedings.neurips.cc/paper_files/paper/2019/file/31ca0ca71184bbdb3de7b20a51e88e90-Paper.pdf}.
\bibitem[{Chaabouni et~al.(2019b)Chaabouni, Kharitonov, Dupoux and Baroni}]{ChaabouniKDB-2019-antiefficient}
\bibinfo{author}{Chaabouni, R.}, \bibinfo{author}{Kharitonov, E.}, \bibinfo{author}{Dupoux, E.}, \bibinfo{author}{Baroni, M.}, \bibinfo{year}{2019}b.
\newblock \bibinfo{title}{Anti-efficient encoding in emergent communication}, in: \bibinfo{booktitle}{Advances in Neural Information Processing Systems 32: Annual Conference on Neural Information Processing Systems 2019, NeurIPS 2019, December 8-14, 2019, Vancouver, BC, Canada}, pp. \bibinfo{pages}{6290--6300}.
\newblock \URLprefix \url{https://proceedings.neurips.cc/paper/2019/hash/31ca0ca71184bbdb3de7b20a51e88e90-Abstract.html}.
\bibitem[{Chaabouni et~al.(2021)Chaabouni, Kharitonov, Dupoux and Baroni}]{ChaabouniKDB-2021-communicating}
\bibinfo{author}{Chaabouni, R.}, \bibinfo{author}{Kharitonov, E.}, \bibinfo{author}{Dupoux, E.}, \bibinfo{author}{Baroni, M.}, \bibinfo{year}{2021}.
\newblock \bibinfo{title}{Communicating artificial neural networks develop efficient color-naming systems}.
\newblock \bibinfo{journal}{Proceedings of the National Academy of Sciences} \bibinfo{volume}{118}, \bibinfo{pages}{e2016569118}.
\newblock \URLprefix \url{https://www.pnas.org/doi/abs/10.1073/pnas.2016569118}, \DOIprefix\doi{10.1073/pnas.2016569118}, \href{http://arxiv.org/abs/https://www.pnas.org/doi/pdf/10.1073/pnas.2016569118}{{\tt arXiv:https://www.pnas.org/doi/pdf/10.1073/pnas.2016569118}}.
\bibitem[{Chandler(2002)}]{Chandler2002}
\bibinfo{author}{Chandler, D.}, \bibinfo{year}{2002}.
\newblock \bibinfo{title}{{Semiotics the Basics}}.
\newblock \bibinfo{publisher}{Routledge}.
\bibitem[{Chu et~al.(2020)Chu, Chinchali and Katti}]{Chu2020Multi-agent}
\bibinfo{author}{Chu, T.}, \bibinfo{author}{Chinchali, S.}, \bibinfo{author}{Katti, S.}, \bibinfo{year}{2020}.
\newblock \bibinfo{title}{Multi-agent reinforcement learning for networked system control}, in: \bibinfo{booktitle}{International Conference on Learning Representations (ICLR)}.
\newblock \URLprefix \url{https://openreview.net/forum?id=Syx7A3NFvH}.
\bibitem[{Constant et~al.(2018)Constant, Ramstead, Veissi{\`e}re, Campbell and Friston}]{Constant2018-hv}
\bibinfo{author}{Constant, A.}, \bibinfo{author}{Ramstead, M.J.D.}, \bibinfo{author}{Veissi{\`e}re, S.P.L.}, \bibinfo{author}{Campbell, J.O.}, \bibinfo{author}{Friston, K.J.}, \bibinfo{year}{2018}.
\newblock \bibinfo{title}{A variational approach to niche construction}.
\newblock \bibinfo{journal}{J. R. Soc. Interface} \bibinfo{volume}{15}.
\bibitem[{Cover and Thomas(2006)}]{Cover2006}
\bibinfo{author}{Cover, T.M.}, \bibinfo{author}{Thomas, J.A.}, \bibinfo{year}{2006}.
\newblock \bibinfo{title}{Elements of Information Theory}.
\newblock \bibinfo{edition}{2nd} ed., \bibinfo{publisher}{Wiley-Interscience}.
\bibitem[{Cowen-Rivers and Naradowsky(2020)}]{cowen2020emergent}
\bibinfo{author}{Cowen-Rivers, A.I.}, \bibinfo{author}{Naradowsky, J.}, \bibinfo{year}{2020}.
\newblock \bibinfo{title}{Emergent communication with world models}.
\newblock \bibinfo{journal}{NeurIPS Workshop on Emergent Communication} .
\bibitem[{Csibra and Gergely(2009)}]{csibra2009natural}
\bibinfo{author}{Csibra, G.}, \bibinfo{author}{Gergely, G.}, \bibinfo{year}{2009}.
\newblock \bibinfo{title}{Natural pedagogy}.
\newblock \bibinfo{journal}{Trends in Cognitive Sciences} \bibinfo{volume}{13}, \bibinfo{pages}{148--153}.
\bibitem[{Deacon(1998)}]{deacon1998symbolic}
\bibinfo{author}{Deacon, T.W.}, \bibinfo{year}{1998}.
\newblock \bibinfo{title}{{The Symbolic Species: The Co-Evolution of Language and the Brain}}.
\newblock \bibinfo{publisher}{W. W. Norton \& Company}.
\bibitem[{Dey et~al.(2024)Dey, Zaech, Nikolov, Van~Gool and Paudel}]{dey2024revla}
\bibinfo{author}{Dey, S.}, \bibinfo{author}{Zaech, J.N.}, \bibinfo{author}{Nikolov, N.}, \bibinfo{author}{Van~Gool, L.}, \bibinfo{author}{Paudel, D.P.}, \bibinfo{year}{2024}.
\newblock \bibinfo{title}{Revla: Reverting visual domain limitation of robotic foundation models}.
\newblock \bibinfo{journal}{arXiv preprint arXiv:2409.15250} .
\bibitem[{Ding et~al.(2024)Ding, Zhang, Shang, Zhang, Zong, Feng, Yuan, Su, Li, Sukiennik et~al.}]{ding2024understanding}
\bibinfo{author}{Ding, J.}, \bibinfo{author}{Zhang, Y.}, \bibinfo{author}{Shang, Y.}, \bibinfo{author}{Zhang, Y.}, \bibinfo{author}{Zong, Z.}, \bibinfo{author}{Feng, J.}, \bibinfo{author}{Yuan, Y.}, \bibinfo{author}{Su, H.}, \bibinfo{author}{Li, N.}, \bibinfo{author}{Sukiennik, N.}, et~al., \bibinfo{year}{2024}.
\newblock \bibinfo{title}{Understanding world or predicting future? a comprehensive survey of world models}.
\newblock \bibinfo{journal}{ACM Computing Surveys} .
\bibitem[{Doya et~al.(2007)Doya, Ishii, Pouget and Rao}]{Doya2007}
\bibinfo{editor}{Doya, K.}, \bibinfo{editor}{Ishii, S.}, \bibinfo{editor}{Pouget, A.}, \bibinfo{editor}{Rao, R.P.N.} (Eds.), \bibinfo{year}{2007}.
\newblock \bibinfo{title}{Bayesian Brain: Probabilistic Approaches to Neural Coding}.
\newblock \bibinfo{publisher}{The MIT Press}.
\bibitem[{Ebara et~al.(2023)Ebara, Nakamura, Taniguchi and Taniguchi}]{ebara2023multi}
\bibinfo{author}{Ebara, H.}, \bibinfo{author}{Nakamura, T.}, \bibinfo{author}{Taniguchi, A.}, \bibinfo{author}{Taniguchi, T.}, \bibinfo{year}{2023}.
\newblock \bibinfo{title}{Multi-agent reinforcement learning with emergent communication using discrete and indifferentiable message}, in: \bibinfo{booktitle}{2023 15th International Congress on Advanced Applied Informatics Winter (IIAI-AAI-Winter)}, pp. \bibinfo{pages}{366--371}.
\bibitem[{Farshbafan et~al.(2022)Farshbafan, Saad and Debbah}]{farshbafan2022common}
\bibinfo{author}{Farshbafan, M.K.}, \bibinfo{author}{Saad, W.}, \bibinfo{author}{Debbah, M.}, \bibinfo{year}{2022}.
\newblock \bibinfo{title}{Common language for goal-oriented semantic communications: A curriculum learning framework}, in: \bibinfo{booktitle}{ICC 2022 - IEEE International Conference on Communications}, \bibinfo{organization}{IEEE}. pp. \bibinfo{pages}{330--335}.
\newblock \DOIprefix\doi{10.1109/ICC45855.2022.9839066}.
\bibitem[{Foerster et~al.(2016a)Foerster, Assael, De~Freitas and Whiteson}]{foerster2016learning}
\bibinfo{author}{Foerster, J.}, \bibinfo{author}{Assael, I.A.}, \bibinfo{author}{De~Freitas, N.}, \bibinfo{author}{Whiteson, S.}, \bibinfo{year}{2016}a.
\newblock \bibinfo{title}{Learning to communicate with deep multi-agent reinforcement learning}.
\newblock \bibinfo{journal}{Advances in Neural Information Processing Systems} \bibinfo{volume}{29}.
\bibitem[{Foerster et~al.(2016b)Foerster, Assael, de~Freitas and Whiteson}]{FoersterAFW-2016-riddles}
\bibinfo{author}{Foerster, J.N.}, \bibinfo{author}{Assael, Y.M.}, \bibinfo{author}{de~Freitas, N.}, \bibinfo{author}{Whiteson, S.}, \bibinfo{year}{2016}b.
\newblock \bibinfo{title}{Learning to communicate to solve riddles with deep distributed recurrent q-networks}.
\newblock \bibinfo{journal}{IJCAI 2016 Deep Learning Workshop} \URLprefix \url{http://arxiv.org/abs/1602.02672}.
\bibitem[{Friston(2010)}]{friston2010free}
\bibinfo{author}{Friston, K.}, \bibinfo{year}{2010}.
\newblock \bibinfo{title}{The free-energy principle: a unified brain theory?}
\newblock \bibinfo{journal}{Nature reviews neuroscience} \bibinfo{volume}{11}, \bibinfo{pages}{127--138}.
\bibitem[{Friston(2013)}]{Friston2013-pf}
\bibinfo{author}{Friston, K.}, \bibinfo{year}{2013}.
\newblock \bibinfo{title}{Life as we know it}.
\newblock \bibinfo{journal}{J. R. Soc. Interface} \bibinfo{volume}{10}, \bibinfo{pages}{20130475}.
\bibitem[{Friston(2019)}]{friston2019free}
\bibinfo{author}{Friston, K.}, \bibinfo{year}{2019}.
\newblock \bibinfo{title}{A free energy principle for a particular physics}.
\newblock \bibinfo{journal}{arXiv preprint arXiv:1906.10184} .
\bibitem[{Friston et~al.(2021)Friston, Moran, Nagai, Taniguchi, Gomi and Tenenbaum}]{friston2021world}
\bibinfo{author}{Friston, K.}, \bibinfo{author}{Moran, R.J.}, \bibinfo{author}{Nagai, Y.}, \bibinfo{author}{Taniguchi, T.}, \bibinfo{author}{Gomi, H.}, \bibinfo{author}{Tenenbaum, J.}, \bibinfo{year}{2021}.
\newblock \bibinfo{title}{World model learning and inference}.
\newblock \bibinfo{journal}{Neural Networks} \bibinfo{volume}{144}, \bibinfo{pages}{573--590}.
\bibitem[{Gurnee and Tegmark(2024)}]{gurnee2023language}
\bibinfo{author}{Gurnee, W.}, \bibinfo{author}{Tegmark, M.}, \bibinfo{year}{2024}.
\newblock \bibinfo{title}{Language models represent space and time}, in: \bibinfo{booktitle}{The Twelfth International Conference on Learning Representations (ICLR)}.
\newblock \URLprefix \url{https://openreview.net/forum?id=jE8xbmvFin}.
\bibitem[{Ha and Schmidhuber(2018a)}]{ha2018recurrent}
\bibinfo{author}{Ha, D.}, \bibinfo{author}{Schmidhuber, J.}, \bibinfo{year}{2018}a.
\newblock \bibinfo{title}{Recurrent world models facilitate policy evolution}, in: \bibinfo{booktitle}{NeurIPS}.
\bibitem[{Ha and Schmidhuber(2018b)}]{ha2018world}
\bibinfo{author}{Ha, D.}, \bibinfo{author}{Schmidhuber, J.}, \bibinfo{year}{2018}b.
\newblock \bibinfo{title}{World models}.
\newblock \bibinfo{journal}{arXiv preprint arXiv:1803.10122} .
\bibitem[{Hafner et~al.(2019a)Hafner, Lillicrap, Ba and Norouzi}]{hafner2019dream}
\bibinfo{author}{Hafner, D.}, \bibinfo{author}{Lillicrap, T.}, \bibinfo{author}{Ba, J.}, \bibinfo{author}{Norouzi, M.}, \bibinfo{year}{2019}a.
\newblock \bibinfo{title}{Dream to control: Learning behaviors by latent imagination}, in: \bibinfo{booktitle}{ICLR}.
\bibitem[{Hafner et~al.(2019b)Hafner, Lillicrap, Fischer, Villegas, Ha, Lee and Davidson}]{hafner2019learning}
\bibinfo{author}{Hafner, D.}, \bibinfo{author}{Lillicrap, T.}, \bibinfo{author}{Fischer, I.}, \bibinfo{author}{Villegas, R.}, \bibinfo{author}{Ha, D.}, \bibinfo{author}{Lee, H.}, \bibinfo{author}{Davidson, J.}, \bibinfo{year}{2019}b.
\newblock \bibinfo{title}{Learning latent dynamics for planning from pixels}, in: \bibinfo{booktitle}{ICML}, \bibinfo{organization}{PMLR}. pp. \bibinfo{pages}{2555--2565}.
\bibitem[{Hafner et~al.(2020)Hafner, Lillicrap, Norouzi and Ba}]{hafner2020mastering}
\bibinfo{author}{Hafner, D.}, \bibinfo{author}{Lillicrap, T.}, \bibinfo{author}{Norouzi, M.}, \bibinfo{author}{Ba, J.}, \bibinfo{year}{2020}.
\newblock \bibinfo{title}{Mastering atari with discrete world models}.
\newblock \bibinfo{journal}{arXiv preprint arXiv:2010.02193} .
\bibitem[{Hafner et~al.(2025)Hafner, Pasukonis, Ba and Lillicrap}]{hafner2025mastering}
\bibinfo{author}{Hafner, D.}, \bibinfo{author}{Pasukonis, J.}, \bibinfo{author}{Ba, J.}, \bibinfo{author}{Lillicrap, T.}, \bibinfo{year}{2025}.
\newblock \bibinfo{title}{Mastering diverse control tasks through world models}.
\newblock \bibinfo{journal}{Nature} , \bibinfo{pages}{1--7}.
\bibitem[{Hagiwara et~al.(2022)Hagiwara, Furukawa, Taniguchi and Taniguchi}]{hagiwara2022multiagent}
\bibinfo{author}{Hagiwara, Y.}, \bibinfo{author}{Furukawa, K.}, \bibinfo{author}{Taniguchi, A.}, \bibinfo{author}{Taniguchi, T.}, \bibinfo{year}{2022}.
\newblock \bibinfo{title}{Multiagent multimodal categorization for symbol emergence: emergent communication via interpersonal cross-modal inference}.
\newblock \bibinfo{journal}{Advanced Robotics} \bibinfo{volume}{36}, \bibinfo{pages}{239--260}.
\bibitem[{Hagiwara et~al.(2019)Hagiwara, Kobayashi, Taniguchi and Taniguchi}]{hagiwara2019symbol}
\bibinfo{author}{Hagiwara, Y.}, \bibinfo{author}{Kobayashi, H.}, \bibinfo{author}{Taniguchi, A.}, \bibinfo{author}{Taniguchi, T.}, \bibinfo{year}{2019}.
\newblock \bibinfo{title}{Symbol emergence as an interpersonal multimodal categorization}.
\newblock \bibinfo{journal}{Frontiers in Robotics and AI} \bibinfo{volume}{6}, \bibinfo{pages}{134}.
\newblock \URLprefix \url{https://www.frontiersin.org/article/10.3389/frobt.2019.00134}, \DOIprefix\doi{10.3389/frobt.2019.00134}.
\bibitem[{Hale(2001)}]{Hale-2001-earley-parser}
\bibinfo{author}{Hale, J.}, \bibinfo{year}{2001}.
\newblock \bibinfo{title}{A probabilistic earley parser as a psycholinguistic model}, in: \bibinfo{booktitle}{Language Technologies 2001: The Second Meeting of the North American Chapter of the Association for Computational Linguistics, {NAACL} 2001, Pittsburgh, PA, USA, June 2-7, 2001}, \bibinfo{publisher}{The Association for Computational Linguistics}.
\newblock \URLprefix \url{https://aclanthology.org/N01-1021/}.
\bibitem[{Hao et~al.(2023)Hao, Gu, Ma, Hong, Wang, Wang and Hu}]{hao2023reasoning}
\bibinfo{author}{Hao, S.}, \bibinfo{author}{Gu, Y.}, \bibinfo{author}{Ma, H.}, \bibinfo{author}{Hong, J.J.}, \bibinfo{author}{Wang, Z.}, \bibinfo{author}{Wang, D.Z.}, \bibinfo{author}{Hu, Z.}, \bibinfo{year}{2023}.
\newblock \bibinfo{title}{Reasoning with language model is planning with world model} \URLprefix \url{https://openreview.net/forum?id=VTWWvYtF1R}.
\bibitem[{Harris(1954)}]{harris1954distributional}
\bibinfo{author}{Harris, Z.}, \bibinfo{year}{1954}.
\newblock \bibinfo{title}{Distributional structure}.
\newblock \bibinfo{journal}{Word} \bibinfo{volume}{10}, \bibinfo{pages}{146--162}.
\bibitem[{Hoang et~al.(2024a)Hoang, Taniguchi, Hagiwara and Taniguchi}]{hoang2024emergent}
\bibinfo{author}{Hoang, N.L.}, \bibinfo{author}{Taniguchi, T.}, \bibinfo{author}{Hagiwara, Y.}, \bibinfo{author}{Taniguchi, A.}, \bibinfo{year}{2024}a.
\newblock \bibinfo{title}{Emergent communication of multimodal deep generative models based on {Metropolis-Hastings} naming game}.
\newblock \bibinfo{journal}{Frontiers in Robotics and AI} \bibinfo{volume}{10}, \bibinfo{pages}{1290604}.
\bibitem[{Hoang et~al.(2024b)Hoang, Taniguchi, Tianwei and Taniguchi}]{hoang2024simsiam}
\bibinfo{author}{Hoang, N.L.}, \bibinfo{author}{Taniguchi, T.}, \bibinfo{author}{Tianwei, F.}, \bibinfo{author}{Taniguchi, A.}, \bibinfo{year}{2024}b.
\newblock \bibinfo{title}{Simsiam naming game: A unified approach for representation learning and emergent communication}.
\newblock \bibinfo{journal}{arXiv preprint arXiv:2410.21803} .
\bibitem[{Hohwy(2013)}]{hohwy2013predictive}
\bibinfo{author}{Hohwy, J.}, \bibinfo{year}{2013}.
\newblock \bibinfo{title}{The predictive mind}.
\newblock \bibinfo{publisher}{Oxford University Press}.
\bibitem[{Huh et~al.(2024)Huh, Cheung, Wang and Isola}]{Huh2024-ax}
\bibinfo{author}{Huh, M.}, \bibinfo{author}{Cheung, B.}, \bibinfo{author}{Wang, T.}, \bibinfo{author}{Isola, P.}, \bibinfo{year}{2024}.
\newblock \bibinfo{title}{Position: The platonic representation hypothesis}, in: \bibinfo{booktitle}{Proceedings of the 41st International Conference on Machine Learning}, \bibinfo{publisher}{PMLR}. pp. \bibinfo{pages}{20617--20642}.
\newblock \URLprefix \url{https://proceedings.mlr.press/v235/huh24a.html}.
\bibitem[{Inukai et~al.(2023)Inukai, Taniguchi, Taniguchi and Hagiwara}]{inukai2023recursive}
\bibinfo{author}{Inukai, J.}, \bibinfo{author}{Taniguchi, T.}, \bibinfo{author}{Taniguchi, A.}, \bibinfo{author}{Hagiwara, Y.}, \bibinfo{year}{2023}.
\newblock \bibinfo{title}{Recursive {Metropolis-Hastings} naming game: Symbol emergence in a multi-agent system based on probabilistic generative models}.
\newblock \bibinfo{journal}{Frontiers in Artificial Intelligence} \bibinfo{volume}{6}.
\bibitem[{Iqbal and Sha(2019)}]{iqbal2019actor}
\bibinfo{author}{Iqbal, S.}, \bibinfo{author}{Sha, F.}, \bibinfo{year}{2019}.
\newblock \bibinfo{title}{Actor-attention-critic for multi-agent reinforcement learning}, in: \bibinfo{booktitle}{International Conference on Machine Learning}, pp. \bibinfo{pages}{2961--2970}.
\bibitem[{Jaques et~al.(2021)Jaques, Burke and Hospedales}]{jaques2021newtonianvae}
\bibinfo{author}{Jaques, M.}, \bibinfo{author}{Burke, M.}, \bibinfo{author}{Hospedales, T.M.}, \bibinfo{year}{2021}.
\newblock \bibinfo{title}{{NewtonianVAE}: Proportional control and goal identification from pixels via physical latent spaces}, in: \bibinfo{booktitle}{CVPR}, pp. \bibinfo{pages}{4454--4463}.
\bibitem[{Jaques et~al.(2019)Jaques, Lazaridou, Hughes, Gulcehre, Ortega, Strouse, Leibo and De~Freitas}]{JaquesLHGOSLF-2019-intrinsic-motivation}
\bibinfo{author}{Jaques, N.}, \bibinfo{author}{Lazaridou, A.}, \bibinfo{author}{Hughes, E.}, \bibinfo{author}{Gulcehre, C.}, \bibinfo{author}{Ortega, P.}, \bibinfo{author}{Strouse, D.}, \bibinfo{author}{Leibo, J.Z.}, \bibinfo{author}{De~Freitas, N.}, \bibinfo{year}{2019}.
\newblock \bibinfo{title}{Social influence as intrinsic motivation for multi-agent deep reinforcement learning}, in: \bibinfo{booktitle}{Proceedings of the 36th International Conference on Machine Learning ({ICML})}, \bibinfo{publisher}{PMLR}. pp. \bibinfo{pages}{3040--3049}.
\newblock \URLprefix \url{https://proceedings.mlr.press/v97/jaques19a.html}.
\bibitem[{Jiang and Lu(2018)}]{jiang2018learning}
\bibinfo{author}{Jiang, J.}, \bibinfo{author}{Lu, Z.}, \bibinfo{year}{2018}.
\newblock \bibinfo{title}{Learning attentional communication for multi-agent cooperation}.
\newblock \bibinfo{journal}{Advances in Neural Information Processing Systems} \bibinfo{volume}{31}.
\bibitem[{Kato et~al.(2024)Kato, Ueda, Naradowsky and Miyao}]{KatoUNM-2024-stack}
\bibinfo{author}{Kato, D.}, \bibinfo{author}{Ueda, R.}, \bibinfo{author}{Naradowsky, J.}, \bibinfo{author}{Miyao, Y.}, \bibinfo{year}{2024}.
\newblock \bibinfo{title}{Emergent communication with stack-based agents}.
\newblock \bibinfo{journal}{Proceedings of the Annual Meeting of the Cognitive Science Society} \bibinfo{volume}{46}.
\bibitem[{Kawaharazuka et~al.(2024)Kawaharazuka, Matsushima, Gambardella, Guo, Paxton and Zeng}]{Kawaharazuka16092024}
\bibinfo{author}{Kawaharazuka, K.}, \bibinfo{author}{Matsushima, T.}, \bibinfo{author}{Gambardella, A.}, \bibinfo{author}{Guo, J.}, \bibinfo{author}{Paxton, C.}, \bibinfo{author}{Zeng, A.}, \bibinfo{year}{2024}.
\newblock \bibinfo{title}{Real-world robot applications of foundation models: a review}.
\newblock \bibinfo{journal}{Advanced Robotics} \bibinfo{volume}{38}, \bibinfo{pages}{1232--1254}.
\newblock \URLprefix \url{https://doi.org/10.1080/01691864.2024.2408593}, \DOIprefix\doi{10.1080/01691864.2024.2408593}, \href{http://arxiv.org/abs/https://doi.org/10.1080/01691864.2024.2408593}{{\tt arXiv:https://doi.org/10.1080/01691864.2024.2408593}}.
\bibitem[{Kilinc and Montana(2018)}]{kilinc2018multi}
\bibinfo{author}{Kilinc, O.}, \bibinfo{author}{Montana, G.}, \bibinfo{year}{2018}.
\newblock \bibinfo{title}{Multi-agent deep reinforcement learning with extremely noisy observations}.
\newblock \bibinfo{journal}{Advances in Neural Information Processing Systems: Deep Reinforcement Learning Workshop} .
\bibitem[{Kim et~al.(2019)Kim, Moon, Hostallero, Kang, Lee, Son and Yi}]{kim2019learning}
\bibinfo{author}{Kim, D.}, \bibinfo{author}{Moon, S.}, \bibinfo{author}{Hostallero, D.}, \bibinfo{author}{Kang, W.J.}, \bibinfo{author}{Lee, T.}, \bibinfo{author}{Son, K.}, \bibinfo{author}{Yi, Y.}, \bibinfo{year}{2019}.
\newblock \bibinfo{title}{Learning to schedule communication in multi-agent reinforcement learning}.
\newblock \bibinfo{journal}{International Conference on Representation Learning} .
\bibitem[{Kim et~al.(2024)Kim, Pertsch, Karamcheti, Xiao, Balakrishna, Nair, Rafailov, Foster, Lam, Sanketi et~al.}]{kim2024openvla}
\bibinfo{author}{Kim, M.J.}, \bibinfo{author}{Pertsch, K.}, \bibinfo{author}{Karamcheti, S.}, \bibinfo{author}{Xiao, T.}, \bibinfo{author}{Balakrishna, A.}, \bibinfo{author}{Nair, S.}, \bibinfo{author}{Rafailov, R.}, \bibinfo{author}{Foster, E.}, \bibinfo{author}{Lam, G.}, \bibinfo{author}{Sanketi, P.}, et~al., \bibinfo{year}{2024}.
\newblock \bibinfo{title}{Openvla: An open-source vision-language-action model}.
\newblock \bibinfo{journal}{8th Conference on Robot Learning (CoRL 2024)} .
\bibitem[{Kingma et~al.(2014)Kingma, Mohamed, Jimenez~Rezende and Welling}]{kingma2014cvae}
\bibinfo{author}{Kingma, D.P.}, \bibinfo{author}{Mohamed, S.}, \bibinfo{author}{Jimenez~Rezende, D.}, \bibinfo{author}{Welling, M.}, \bibinfo{year}{2014}.
\newblock \bibinfo{title}{Semi-supervised learning with deep generative models}, in: \bibinfo{editor}{Ghahramani, Z.}, \bibinfo{editor}{Welling, M.}, \bibinfo{editor}{Cortes, C.}, \bibinfo{editor}{Lawrence, N.}, \bibinfo{editor}{Weinberger, K.} (Eds.), \bibinfo{booktitle}{Advances in Neural Information Processing Systems}, \bibinfo{publisher}{Curran Associates, Inc.}
\newblock \URLprefix \url{https://proceedings.neurips.cc/paper_files/paper/2014/file/d523773c6b194f37b938d340d5d02232-Paper.pdf}.
\bibitem[{Kinose et~al.(2023)Kinose, Okada, Okumura and Taniguchi}]{kinose2022multi}
\bibinfo{author}{Kinose, A.}, \bibinfo{author}{Okada, M.}, \bibinfo{author}{Okumura, R.}, \bibinfo{author}{Taniguchi, T.}, \bibinfo{year}{2023}.
\newblock \bibinfo{title}{Multi-view dreaming: Multi-view world model with contrastive learning}.
\newblock \bibinfo{journal}{Advanced Robotics} \bibinfo{volume}{37}, \bibinfo{pages}{1212--1220}.
\bibitem[{Kirchhoff et~al.(2018)Kirchhoff, Parr, Palacios, Friston and Kiverstein}]{Kirchhoff2018-ma}
\bibinfo{author}{Kirchhoff, M.}, \bibinfo{author}{Parr, T.}, \bibinfo{author}{Palacios, E.}, \bibinfo{author}{Friston, K.}, \bibinfo{author}{Kiverstein, J.}, \bibinfo{year}{2018}.
\newblock \bibinfo{title}{The markov blankets of life: autonomy, active inference and the free energy principle}.
\newblock \bibinfo{journal}{J. R. Soc. Interface} \bibinfo{volume}{15}.
\bibitem[{Kottur et~al.(2017)Kottur, Moura, Lee and Batra}]{KotturMLB-2017-compositionality}
\bibinfo{author}{Kottur, S.}, \bibinfo{author}{Moura, J.M.F.}, \bibinfo{author}{Lee, S.}, \bibinfo{author}{Batra, D.}, \bibinfo{year}{2017}.
\newblock \bibinfo{title}{Natural language does not emerge 'naturally' in multi-agent dialog}, in: \bibinfo{editor}{Palmer, M.}, \bibinfo{editor}{Hwa, R.}, \bibinfo{editor}{Riedel, S.} (Eds.), \bibinfo{booktitle}{Proceedings of the 2017 Conference on Empirical Methods in Natural Language Processing, {EMNLP} 2017, Copenhagen, Denmark, September 9-11, 2017}, \bibinfo{publisher}{Association for Computational Linguistics}. pp. \bibinfo{pages}{2962--2967}.
\newblock \URLprefix \url{https://doi.org/10.18653/v1/d17-1321}, \DOIprefix\doi{10.18653/V1/D17-1321}.
\bibitem[{Kuribayashi et~al.(2022)Kuribayashi, Oseki, Brassard and Inui}]{KuribayashiOBI-2022-context}
\bibinfo{author}{Kuribayashi, T.}, \bibinfo{author}{Oseki, Y.}, \bibinfo{author}{Brassard, A.}, \bibinfo{author}{Inui, K.}, \bibinfo{year}{2022}.
\newblock \bibinfo{title}{Context limitations make neural language models more human-like}, in: \bibinfo{editor}{Goldberg, Y.}, \bibinfo{editor}{Kozareva, Z.}, \bibinfo{editor}{Zhang, Y.} (Eds.), \bibinfo{booktitle}{Proceedings of the 2022 Conference on Empirical Methods in Natural Language Processing, {EMNLP} 2022, Abu Dhabi, United Arab Emirates, December 7-11, 2022}, \bibinfo{publisher}{Association for Computational Linguistics}. pp. \bibinfo{pages}{10421--10436}.
\newblock \URLprefix \url{https://doi.org/10.18653/v1/2022.emnlp-main.712}, \DOIprefix\doi{10.18653/V1/2022.EMNLP-MAIN.712}.
\bibitem[{Laskin et~al.(2020)Laskin, Srinivas and Abbeel}]{laskin2020curl}
\bibinfo{author}{Laskin, M.}, \bibinfo{author}{Srinivas, A.}, \bibinfo{author}{Abbeel, P.}, \bibinfo{year}{2020}.
\newblock \bibinfo{title}{{CURL}: Contrastive unsupervised representations for reinforcement learning}, in: \bibinfo{booktitle}{ICML}, \bibinfo{organization}{PMLR}. pp. \bibinfo{pages}{5639--5650}.
\bibitem[{Lazaridou and Baroni(2020)}]{LazaridouB-2020-emergent}
\bibinfo{author}{Lazaridou, A.}, \bibinfo{author}{Baroni, M.}, \bibinfo{year}{2020}.
\newblock \bibinfo{title}{Emergent multi-agent communication in the deep learning era}.
\newblock \bibinfo{journal}{arXiv preprint arXiv:2006.02419} \URLprefix \url{https://arxiv.org/abs/2006.02419}, \href{http://arxiv.org/abs/2006.02419}{{\tt arXiv:2006.02419}}.
\bibitem[{Le~Hoang et~al.(2024)Le~Hoang, Matsui, Hagiwara, Taniguchi and Taniguchi}]{hoang2024compositionality}
\bibinfo{author}{Le~Hoang, N.}, \bibinfo{author}{Matsui, Y.}, \bibinfo{author}{Hagiwara, Y.}, \bibinfo{author}{Taniguchi, A.}, \bibinfo{author}{Taniguchi, T.}, \bibinfo{year}{2024}.
\newblock \bibinfo{title}{Compositionality and generalization in emergent communication using {Metropolis-Hastings} naming game}, in: \bibinfo{booktitle}{2024 IEEE International Conference on Development and Learning (ICDL)}, \bibinfo{organization}{IEEE}. pp. \bibinfo{pages}{1--7}.
\bibitem[{LeCun(2022)}]{lecun2022path}
\bibinfo{author}{LeCun, Y.}, \bibinfo{year}{2022}.
\newblock \bibinfo{title}{A path towards autonomous machine intelligence version 0.9. 2, 2022-06-27}.
\newblock \bibinfo{journal}{OpenReview} \URLprefix \url{https://openreview.net/forum?id=BZ5a1r-kVsf}.
\bibitem[{Levine(2018)}]{DBLP:journals/corr/abs-1805-00909}
\bibinfo{author}{Levine, S.}, \bibinfo{year}{2018}.
\newblock \bibinfo{title}{Reinforcement learning and control as probabilistic inference: Tutorial and review}.
\newblock \bibinfo{journal}{arXiv:1805.00909} .
\bibitem[{Levy(2008)}]{Levy-2008-expectation-based}
\bibinfo{author}{Levy, R.}, \bibinfo{year}{2008}.
\newblock \bibinfo{title}{Expectation-based syntactic comprehension}.
\newblock \bibinfo{journal}{Cognition} \bibinfo{volume}{106}, \bibinfo{pages}{1126--1177}.
\newblock \URLprefix \url{https://www.sciencedirect.com/science/article/pii/S0010027707001436}, \DOIprefix\doi{https://doi.org/10.1016/j.cognition.2007.05.006}.
\bibitem[{Lewis(1969)}]{Lewis-1969-convention}
\bibinfo{author}{Lewis, D.K.}, \bibinfo{year}{1969}.
\newblock \bibinfo{title}{Convention: A Philosophical Study}.
\newblock \bibinfo{publisher}{Wiley-Blackwell}.
\bibitem[{Li et~al.(2023)Li, Hopkins, Bau, Vi{\'e}gas, Pfister and Wattenberg}]{li2022emergent}
\bibinfo{author}{Li, K.}, \bibinfo{author}{Hopkins, A.K.}, \bibinfo{author}{Bau, D.}, \bibinfo{author}{Vi{\'e}gas, F.}, \bibinfo{author}{Pfister, H.}, \bibinfo{author}{Wattenberg, M.}, \bibinfo{year}{2023}.
\newblock \bibinfo{title}{Emergent world representations: Exploring a sequence model trained on a synthetic task}, in: \bibinfo{booktitle}{The Eleventh International Conference on Learning Representations (ICLR)}.
\newblock \URLprefix \url{https://openreview.net/forum?id=DeG07_TcZvT}.
\bibitem[{Lillicrap et~al.(2015)Lillicrap, Hunt, Pritzel, Heess, Erez, Tassa, Silver and Wierstra}]{lillicrap2015continuous}
\bibinfo{author}{Lillicrap, T.P.}, \bibinfo{author}{Hunt, J.J.}, \bibinfo{author}{Pritzel, A.}, \bibinfo{author}{Heess, N.}, \bibinfo{author}{Erez, T.}, \bibinfo{author}{Tassa, Y.}, \bibinfo{author}{Silver, D.}, \bibinfo{author}{Wierstra, D.}, \bibinfo{year}{2015}.
\newblock \bibinfo{title}{Continuous control with deep reinforcement learning}.
\newblock \bibinfo{journal}{arXiv preprint arXiv:1509.02971} .
\bibitem[{Liu et~al.(2020)Liu, Wang, Hu, Hao, Chen and Gao}]{liu2020multi}
\bibinfo{author}{Liu, Y.}, \bibinfo{author}{Wang, W.}, \bibinfo{author}{Hu, Y.}, \bibinfo{author}{Hao, J.}, \bibinfo{author}{Chen, X.}, \bibinfo{author}{Gao, Y.}, \bibinfo{year}{2020}.
\newblock \bibinfo{title}{Multi-agent game abstraction via graph attention neural network}, in: \bibinfo{booktitle}{Proceedings of the AAAI Conference on Artificial Intelligence}, pp. \bibinfo{pages}{7211--7218}.
\bibitem[{Lobos-Tsunekawa et~al.(2022)Lobos-Tsunekawa, Srinivasan and Spranger}]{lobos2022ma}
\bibinfo{author}{Lobos-Tsunekawa, K.}, \bibinfo{author}{Srinivasan, A.}, \bibinfo{author}{Spranger, M.}, \bibinfo{year}{2022}.
\newblock \bibinfo{title}{Ma-dreamer: Coordination and communication through shared imagination}.
\newblock \bibinfo{journal}{arXiv preprint arXiv:2204.04687} .
\bibitem[{Lowe et~al.(2017)Lowe, Wu, Tamar, Harb, Pieter~Abbeel and Mordatch}]{lowe2017multi}
\bibinfo{author}{Lowe, R.}, \bibinfo{author}{Wu, Y.I.}, \bibinfo{author}{Tamar, A.}, \bibinfo{author}{Harb, J.}, \bibinfo{author}{Pieter~Abbeel, O.}, \bibinfo{author}{Mordatch, I.}, \bibinfo{year}{2017}.
\newblock \bibinfo{title}{Multi-agent actor-critic for mixed cooperative-competitive environments}.
\newblock \bibinfo{journal}{Advances in Neural Information Processing Systems} \bibinfo{volume}{30}.
\bibitem[{Mahowald et~al.(2024)Mahowald, Ivanova, Blank, Kanwisher, Tenenbaum and Fedorenko}]{mahowald2023dissociating}
\bibinfo{author}{Mahowald, K.}, \bibinfo{author}{Ivanova, A.A.}, \bibinfo{author}{Blank, I.A.}, \bibinfo{author}{Kanwisher, N.}, \bibinfo{author}{Tenenbaum, J.B.}, \bibinfo{author}{Fedorenko, E.}, \bibinfo{year}{2024}.
\newblock \bibinfo{title}{Dissociating language and thought in large language models}.
\newblock \bibinfo{journal}{Trends in Cognitive Sciences} \bibinfo{volume}{28}, \bibinfo{pages}{517--540}.
\newblock \URLprefix \url{https://www.sciencedirect.com/science/article/pii/S1364661324000275}, \DOIprefix\doi{https://doi.org/10.1016/j.tics.2024.01.011}.
\bibitem[{Matsui et~al.(2025)Matsui, Yamaki, Ueda, Shinagawa and Taniguchi}]{matsui2025metropolis}
\bibinfo{author}{Matsui, Y.}, \bibinfo{author}{Yamaki, R.}, \bibinfo{author}{Ueda, R.}, \bibinfo{author}{Shinagawa, S.}, \bibinfo{author}{Taniguchi, T.}, \bibinfo{year}{2025}.
\newblock \bibinfo{title}{Metropolis-hastings captioning game: Knowledge fusion of vision language models via decentralized bayesian inference}.
\newblock \bibinfo{journal}{arXiv preprint arXiv:2504.09620} .
\bibitem[{Mikolov et~al.(2013a)Mikolov, Corrado, Chen and Dean}]{Mikolov2013a}
\bibinfo{author}{Mikolov, T.}, \bibinfo{author}{Corrado, G.}, \bibinfo{author}{Chen, K.}, \bibinfo{author}{Dean, J.}, \bibinfo{year}{2013}a.
\newblock \bibinfo{title}{{Efficient Estimation of Word Representations in Vector Space}}, in: \bibinfo{booktitle}{{International Conference on Learning Representations (ICLR)}}, pp. \bibinfo{pages}{1--12}.
\bibitem[{Mikolov et~al.(2013b)Mikolov, Sutskever, Chen, Corrado and Dean}]{Mikolov2013}
\bibinfo{author}{Mikolov, T.}, \bibinfo{author}{Sutskever, I.}, \bibinfo{author}{Chen, K.}, \bibinfo{author}{Corrado, G.S.}, \bibinfo{author}{Dean, J.}, \bibinfo{year}{2013}b.
\newblock \bibinfo{title}{Distributed representations of words and phrases and their compositionality}, in: \bibinfo{booktitle}{Advances in Neural Information Processing Systems (NeurIPS)}, pp. \bibinfo{pages}{3111--3119}.
\newblock \URLprefix \url{https://proceedings.neurips.cc/paper_files/paper/2013/file/9aa42b31882ec039965f3c4923ce901b-Paper.pdf}.
\bibitem[{Min et~al.(2023)Min, Ross, Sulem, Veyseh, Nguyen, Sainz, Agirre, Heintz and Roth}]{Min2023-dp}
\bibinfo{author}{Min, B.}, \bibinfo{author}{Ross, H.}, \bibinfo{author}{Sulem, E.}, \bibinfo{author}{Veyseh, A.P.B.}, \bibinfo{author}{Nguyen, T.H.}, \bibinfo{author}{Sainz, O.}, \bibinfo{author}{Agirre, E.}, \bibinfo{author}{Heintz, I.}, \bibinfo{author}{Roth, D.}, \bibinfo{year}{2023}.
\newblock \bibinfo{title}{Recent advances in natural language processing via large pre-trained language models: A survey}.
\newblock \bibinfo{journal}{ACM Comput. Surv.} \bibinfo{volume}{56}, \bibinfo{pages}{1--40}.
\bibitem[{Mordatch and Abbeel(2018)}]{MordatchA-2018-emergence}
\bibinfo{author}{Mordatch, I.}, \bibinfo{author}{Abbeel, P.}, \bibinfo{year}{2018}.
\newblock \bibinfo{title}{Emergence of grounded compositional language in multi-agent populations}, in: \bibinfo{booktitle}{Proceedings of the AAAI conference on artificial intelligence}, \bibinfo{publisher}{{AAAI} Press}. pp. \bibinfo{pages}{1495--1502}.
\newblock \URLprefix \url{https://doi.org/10.1609/aaai.v32i1.11492}, \DOIprefix\doi{10.1609/AAAI.V32I1.11492}.
\bibitem[{Nakamura et~al.(2023)Nakamura, Taniguchi and Taniguchi}]{nakamura2023controlprobabilisticinferenceemergent}
\bibinfo{author}{Nakamura, T.}, \bibinfo{author}{Taniguchi, A.}, \bibinfo{author}{Taniguchi, T.}, \bibinfo{year}{2023}.
\newblock \bibinfo{title}{Control as probabilistic inference as an emergent communication mechanism in multi-agent reinforcement learning}.
\newblock \URLprefix \url{https://arxiv.org/abs/2307.05004}, \href{http://arxiv.org/abs/2307.05004}{{\tt arXiv:2307.05004}}.
\bibitem[{Nilsson et~al.(1984)}]{nilsson1984shakey}
\bibinfo{author}{Nilsson, N.J.}, et~al., \bibinfo{year}{1984}.
\newblock \bibinfo{title}{Shakey the robot}. volume \bibinfo{volume}{323}.
\newblock \bibinfo{publisher}{Sri International Menlo Park, California}.
\bibitem[{Niu et~al.(2021)Niu, Paleja and Gombolay}]{niu2021multi}
\bibinfo{author}{Niu, Y.}, \bibinfo{author}{Paleja, R.R.}, \bibinfo{author}{Gombolay, M.C.}, \bibinfo{year}{2021}.
\newblock \bibinfo{title}{Multi-agent graph-attention communication and teaming.}, in: \bibinfo{booktitle}{AAMAS}, pp. \bibinfo{pages}{964--973}.
\bibitem[{Nomura et~al.(2025)Nomura, Aoki, Taniguchi and Horii}]{nomura2025decentralized}
\bibinfo{author}{Nomura, K.}, \bibinfo{author}{Aoki, T.}, \bibinfo{author}{Taniguchi, T.}, \bibinfo{author}{Horii, T.}, \bibinfo{year}{2025}.
\newblock \bibinfo{title}{Decentralized collective world model for emergent communication and coordination}.
\newblock \bibinfo{journal}{arXiv preprint arXiv:2504.03353} .
\bibitem[{Okada and Taniguchi(2021)}]{okada2021dreaming}
\bibinfo{author}{Okada, M.}, \bibinfo{author}{Taniguchi, T.}, \bibinfo{year}{2021}.
\newblock \bibinfo{title}{Dreaming: Model-based reinforcement learning by latent imagination without reconstruction}, in: \bibinfo{booktitle}{ICRA}, \bibinfo{organization}{IEEE}. pp. \bibinfo{pages}{4209--4215}.
\bibitem[{Okada and Taniguchi(2022)}]{okada2022dreamingv2}
\bibinfo{author}{Okada, M.}, \bibinfo{author}{Taniguchi, T.}, \bibinfo{year}{2022}.
\newblock \bibinfo{title}{Dreamingv2: Reinforcement learning with discrete world models without reconstruction}, in: \bibinfo{booktitle}{IROS}.
\bibitem[{Okumura et~al.(2022)Okumura, Nishio and Taniguchi}]{okumura2022tactile}
\bibinfo{author}{Okumura, R.}, \bibinfo{author}{Nishio, N.}, \bibinfo{author}{Taniguchi, T.}, \bibinfo{year}{2022}.
\newblock \bibinfo{title}{{Tactile-Sensitive NewtonianVAE} for high-accuracy industrial connector-socket insertion}, in: \bibinfo{booktitle}{IROS}.
\bibitem[{Okumura et~al.(2023)Okumura, Taniguchi, Hagiwara and Taniguchi}]{okumura2023metropolishastings}
\bibinfo{author}{Okumura, R.}, \bibinfo{author}{Taniguchi, T.}, \bibinfo{author}{Hagiwara, Y.}, \bibinfo{author}{Taniguchi, A.}, \bibinfo{year}{2023}.
\newblock \bibinfo{title}{{Metropolis-Hastings} algorithm in joint-attention naming game: experimental semiotics study}.
\newblock \bibinfo{journal}{Frontiers in Artificial Intelligence} \bibinfo{volume}{6}.
\newblock \URLprefix \url{https://www.frontiersin.org/articles/10.3389/frai.2023.1235231}, \DOIprefix\doi{10.3389/frai.2023.1235231}.
\bibitem[{Osada et~al.(2024)Osada, Garcia~Ricardez, Suzuki and Taniguchi}]{osada2024reflectance}
\bibinfo{author}{Osada, M.}, \bibinfo{author}{Garcia~Ricardez, G.A.}, \bibinfo{author}{Suzuki, Y.}, \bibinfo{author}{Taniguchi, T.}, \bibinfo{year}{2024}.
\newblock \bibinfo{title}{Reflectance estimation for proximity sensing by vision-language models: Utilizing distributional semantics for low-level cognition in robotics}.
\newblock \bibinfo{journal}{Advanced Robotics} \bibinfo{volume}{38}, \bibinfo{pages}{1287--1306}.
\bibitem[{Parr et~al.(2022)Parr, Pezzulo and Friston}]{Parr_Thomas2022-03-29}
\bibinfo{author}{Parr, T.}, \bibinfo{author}{Pezzulo, G.}, \bibinfo{author}{Friston, K.J.}, \bibinfo{year}{2022}.
\newblock \bibinfo{title}{Active Inference: The Free Energy Principle in Mind, Brain, and Behavior (English Edition)}.
\newblock \bibinfo{publisher}{The MIT Press}.
\newblock \URLprefix \url{https://lead.to/amazon/jp/?op=bt&la=ja&key=B096DBD7GX}.
\bibitem[{Peirce(1974)}]{peirce1974collected}
\bibinfo{author}{Peirce, C.S.}, \bibinfo{year}{1974}.
\newblock \bibinfo{title}{Collected papers of charles sanders peirce}. volume~\bibinfo{volume}{5}.
\newblock \bibinfo{publisher}{Harvard University Press}.
\bibitem[{Peirce(1991)}]{peirce1991peirce}
\bibinfo{author}{Peirce, C.S.}, \bibinfo{year}{1991}.
\newblock \bibinfo{title}{Peirce on signs: Writings on semiotic}.
\newblock \bibinfo{publisher}{UNC Press Books}.
\bibitem[{Peters et~al.(2025)Peters, Waubert~de Puiseau, Tercan, Gopikrishnan, Bitter, de~Carvalho and Meisen}]{peters2025emergent}
\bibinfo{author}{Peters, J.}, \bibinfo{author}{Waubert~de Puiseau, C.}, \bibinfo{author}{Tercan, H.}, \bibinfo{author}{Gopikrishnan, A.}, \bibinfo{author}{Bitter, C.}, \bibinfo{author}{de~Carvalho, G.A.L.}, \bibinfo{author}{Meisen, T.}, \bibinfo{year}{2025}.
\newblock \bibinfo{title}{Emergent language: a survey and taxonomy}.
\newblock \bibinfo{journal}{Autonomous Agents and Multi-Agent Systems} \bibinfo{volume}{39}, \bibinfo{pages}{18}.
\newblock \DOIprefix\doi{10.1007/s10458-025-09691-y}.
\bibitem[{Peters et~al.(2024)Peters, de~Puiseau, Tercan, Gopikrishnan, De~Carvalho, Bitter and Meisen}]{peters2024survey}
\bibinfo{author}{Peters, J.}, \bibinfo{author}{de~Puiseau, C.W.}, \bibinfo{author}{Tercan, H.}, \bibinfo{author}{Gopikrishnan, A.}, \bibinfo{author}{De~Carvalho, G.A.L.}, \bibinfo{author}{Bitter, C.}, \bibinfo{author}{Meisen, T.}, \bibinfo{year}{2024}.
\newblock \bibinfo{title}{A survey on emergent language}.
\newblock \bibinfo{journal}{arXiv preprint arXiv:2409.02645} .
\bibitem[{Poole et~al.(2019)Poole, Ozair, van~den Oord, Alemi and Tucker}]{PooleOOAT-2019-on}
\bibinfo{author}{Poole, B.}, \bibinfo{author}{Ozair, S.}, \bibinfo{author}{van~den Oord, A.}, \bibinfo{author}{Alemi, A.A.}, \bibinfo{author}{Tucker, G.}, \bibinfo{year}{2019}.
\newblock \bibinfo{title}{On variational bounds of mutual information}, in: \bibinfo{editor}{Chaudhuri, K.}, \bibinfo{editor}{Salakhutdinov, R.} (Eds.), \bibinfo{booktitle}{Proceedings of the 36th International Conference on Machine Learning, {ICML} 2019, 9-15 June 2019, Long Beach, California, {USA}}, \bibinfo{publisher}{{PMLR}}. pp. \bibinfo{pages}{5171--5180}.
\newblock \URLprefix \url{http://proceedings.mlr.press/v97/poole19a.html}.
\bibitem[{Popovski et~al.(2020)Popovski, Simeone, Boccardi, G{\"u}nd{\"u}z and Sahin}]{popovski2020semantic}
\bibinfo{author}{Popovski, P.}, \bibinfo{author}{Simeone, O.}, \bibinfo{author}{Boccardi, F.}, \bibinfo{author}{G{\"u}nd{\"u}z, D.}, \bibinfo{author}{Sahin, O.}, \bibinfo{year}{2020}.
\newblock \bibinfo{title}{Semantic-effectiveness filtering and control for post-5g wireless connectivity}.
\newblock \bibinfo{journal}{Journal of the Indian Institute of Science} \bibinfo{volume}{100}, \bibinfo{pages}{435--443}.
\bibitem[{Qu et~al.(2020)Qu, Li, Liu, Xiong, Chu, Wang, Qi, Song et~al.}]{qu2020intention}
\bibinfo{author}{Qu, C.}, \bibinfo{author}{Li, H.}, \bibinfo{author}{Liu, C.}, \bibinfo{author}{Xiong, J.}, \bibinfo{author}{Chu, W.}, \bibinfo{author}{Wang, W.}, \bibinfo{author}{Qi, Y.}, \bibinfo{author}{Song, L.}, et~al., \bibinfo{year}{2020}.
\newblock \bibinfo{title}{Intention propagation for multi-agent reinforcement learning} .
\bibitem[{Ramesh et~al.(2021)Ramesh, Pavlov, Goh, Gray, Voss, Radford, Chen and Sutskever}]{ramesh2021zero}
\bibinfo{author}{Ramesh, A.}, \bibinfo{author}{Pavlov, M.}, \bibinfo{author}{Goh, G.}, \bibinfo{author}{Gray, S.}, \bibinfo{author}{Voss, C.}, \bibinfo{author}{Radford, A.}, \bibinfo{author}{Chen, M.}, \bibinfo{author}{Sutskever, I.}, \bibinfo{year}{2021}.
\newblock \bibinfo{title}{Zero-shot text-to-image generation}.
\newblock \bibinfo{journal}{arXiv preprint arXiv:2102.12092} .
\bibitem[{Ren et~al.(2020)Ren, Guo, Labeau, Cohen and Kirby}]{RenGLCK-2020-compositional}
\bibinfo{author}{Ren, Y.}, \bibinfo{author}{Guo, S.}, \bibinfo{author}{Labeau, M.}, \bibinfo{author}{Cohen, S.B.}, \bibinfo{author}{Kirby, S.}, \bibinfo{year}{2020}.
\newblock \bibinfo{title}{Compositional languages emerge in a neural iterated learning model}, in: \bibinfo{booktitle}{8th International Conference on Learning Representations, {ICLR} 2020, Addis Ababa, Ethiopia, April 26-30, 2020}, \bibinfo{publisher}{OpenReview.net}.
\newblock \URLprefix \url{https://openreview.net/forum?id=HkePNpVKPB}.
\bibitem[{Ri et~al.(2023)Ri, Ueda and Naradowsky}]{RiUN-2023-attention}
\bibinfo{author}{Ri, R.}, \bibinfo{author}{Ueda, R.}, \bibinfo{author}{Naradowsky, J.}, \bibinfo{year}{2023}.
\newblock \bibinfo{title}{Emergent communication with attention}, in: \bibinfo{editor}{Goldwater, M.B.}, \bibinfo{editor}{Anggoro, F.K.}, \bibinfo{editor}{Hayes, B.K.}, \bibinfo{editor}{Ong, D.C.} (Eds.), \bibinfo{booktitle}{Proceedings of the 45th Annual Meeting of the Cognitive Science Society, CogSci 2023, Sydney, NSW, Australia, July 26-29, 2023}, \bibinfo{publisher}{cognitivesciencesociety.org}.
\newblock \URLprefix \url{https://escholarship.org/uc/item/7dg8r8zk}.
\bibitem[{Rita et~al.(2020)Rita, Chaabouni and Dupoux}]{RitaCD-2020-lazimpa}
\bibinfo{author}{Rita, M.}, \bibinfo{author}{Chaabouni, R.}, \bibinfo{author}{Dupoux, E.}, \bibinfo{year}{2020}.
\newblock \bibinfo{title}{"lazimpa": Lazy and impatient neural agents learn to communicate efficiently}, in: \bibinfo{booktitle}{Proceedings of the 24th Conference on Computational Natural Language Learning, CoNLL 2020, Online, November 19-20, 2020}, \bibinfo{publisher}{Association for Computational Linguistics}. pp. \bibinfo{pages}{335--343}.
\newblock \URLprefix \url{https://doi.org/10.18653/v1/2020.conll-1.26}, \DOIprefix\doi{10.18653/v1/2020.conll-1.26}.
\bibitem[{Rita et~al.(2022)Rita, Tallec, Michel, Grill, Pietquin, Dupoux and Strub}]{RitaTMGPDS-2022-emergent}
\bibinfo{author}{Rita, M.}, \bibinfo{author}{Tallec, C.}, \bibinfo{author}{Michel, P.}, \bibinfo{author}{Grill, J.}, \bibinfo{author}{Pietquin, O.}, \bibinfo{author}{Dupoux, E.}, \bibinfo{author}{Strub, F.}, \bibinfo{year}{2022}.
\newblock \bibinfo{title}{Emergent communication: Generalization and overfitting in lewis games}, in: \bibinfo{editor}{Koyejo, S.}, \bibinfo{editor}{Mohamed, S.}, \bibinfo{editor}{Agarwal, A.}, \bibinfo{editor}{Belgrave, D.}, \bibinfo{editor}{Cho, K.}, \bibinfo{editor}{Oh, A.} (Eds.), \bibinfo{booktitle}{Advances in Neural Information Processing Systems 35: Annual Conference on Neural Information Processing Systems 2022, NeurIPS 2022, New Orleans, LA, USA, November 28 - December 9, 2022}.
\bibitem[{Saito et~al.(2024)Saito, Nakamura, Taniguchi, Taniguchi, Hayamizu and Zhang}]{saito2024emergence}
\bibinfo{author}{Saito, I.}, \bibinfo{author}{Nakamura, T.}, \bibinfo{author}{Taniguchi, A.}, \bibinfo{author}{Taniguchi, T.}, \bibinfo{author}{Hayamizu, Y.}, \bibinfo{author}{Zhang, S.}, \bibinfo{year}{2024}.
\newblock \bibinfo{title}{Emergence of continuous signals as shared symbols through emergent communication}, in: \bibinfo{booktitle}{2024 IEEE International Conference on Development and Learning (ICDL)}, \bibinfo{organization}{IEEE}. pp. \bibinfo{pages}{1--6}.
\bibitem[{Schmidhuber(1990)}]{schmidhuber1990making}
\bibinfo{author}{Schmidhuber, J.}, \bibinfo{year}{1990}.
\newblock \bibinfo{title}{Making the world differentiable: on using self supervised fully recurrent neural networks for dynamic reinforcement learning and planning in non-stationary environments}. volume \bibinfo{volume}{126}.
\newblock \bibinfo{publisher}{Inst. f{\"u}r Informatik}.
\bibitem[{Seo et~al.(2022)Seo, Park, Bennis and Debbah}]{seo2021semantics}
\bibinfo{author}{Seo, H.E.}, \bibinfo{author}{Park, J.}, \bibinfo{author}{Bennis, M.}, \bibinfo{author}{Debbah, M.}, \bibinfo{year}{2022}.
\newblock \bibinfo{title}{Semantics-native communication with contextual reasoning}.
\newblock \bibinfo{journal}{IEEE Journal on Selected Areas in Communications} \bibinfo{volume}{40}, \bibinfo{pages}{2545--2559}.
\newblock \DOIprefix\doi{10.1109/JSAC.2022.3191299}.
\bibitem[{Shannon(1948)}]{shannon1948mathematical}
\bibinfo{author}{Shannon, C.E.}, \bibinfo{year}{1948}.
\newblock \bibinfo{title}{A mathematical theory of communication}.
\newblock \bibinfo{journal}{The Bell system technical journal} \bibinfo{volume}{27}, \bibinfo{pages}{379--423}.
\bibitem[{Sohn et~al.(2015)Sohn, Lee and Yan}]{sohn2015cvae}
\bibinfo{author}{Sohn, K.}, \bibinfo{author}{Lee, H.}, \bibinfo{author}{Yan, X.}, \bibinfo{year}{2015}.
\newblock \bibinfo{title}{Learning structured output representation using deep conditional generative models}, in: \bibinfo{editor}{Cortes, C.}, \bibinfo{editor}{Lawrence, N.D.}, \bibinfo{editor}{Lee, D.D.}, \bibinfo{editor}{Sugiyama, M.}, \bibinfo{editor}{Garnett, R.} (Eds.), \bibinfo{booktitle}{Advances in Neural Information Processing Systems 28: Annual Conference on Neural Information Processing Systems 2015, December 7-12, 2015, Montreal, Quebec, Canada}, pp. \bibinfo{pages}{3483--3491}.
\newblock \URLprefix \url{https://proceedings.neurips.cc/paper/2015/hash/8d55a249e6baa5c06772297520da2051-Abstract.html}.
\bibitem[{Steels(1997)}]{Steels97}
\bibinfo{author}{Steels, L.}, \bibinfo{year}{1997}.
\newblock \bibinfo{title}{The synthetic modeling of language origins}.
\newblock \bibinfo{journal}{Evolution of Communication Journal} \bibinfo{volume}{1}, \bibinfo{pages}{1--34}.
\newblock \DOIprefix\doi{10.1075/eoc.1.1.02ste}.
\bibitem[{Steels(2005)}]{Steels2007}
\bibinfo{author}{Steels, L.}, \bibinfo{year}{2005}.
\newblock \bibinfo{title}{The emergence and evolution of linguistic structure: from lexical to grammatical communication systems}.
\newblock \bibinfo{journal}{Connection Science} \bibinfo{volume}{17}, \bibinfo{pages}{213--230}.
\newblock \DOIprefix\doi{10.1080/09540090500269088}.
\bibitem[{Steels(2011)}]{STEELS2011339}
\bibinfo{author}{Steels, L.}, \bibinfo{year}{2011}.
\newblock \bibinfo{title}{{Modeling the Cultural Evolution of Language}}.
\newblock \bibinfo{journal}{Physics of Life Reviews} \bibinfo{volume}{8}, \bibinfo{pages}{339--356}.
\newblock \URLprefix \url{https://www.sciencedirect.com/science/article/pii/S1571064511001060}, \DOIprefix\doi{https://doi.org/10.1016/j.plrev.2011.10.014}.
\bibitem[{Sukhbaatar et~al.(2016)Sukhbaatar, Fergus et~al.}]{sukhbaatar2016learning}
\bibinfo{author}{Sukhbaatar, S.}, \bibinfo{author}{Fergus, R.}, et~al., \bibinfo{year}{2016}.
\newblock \bibinfo{title}{Learning multiagent communication with backpropagation}.
\newblock \bibinfo{journal}{Advances in Neural Information Processing Systems} \bibinfo{volume}{29}.
\bibitem[{Sutton(1990)}]{sutton1990integrated}
\bibinfo{author}{Sutton, R.S.}, \bibinfo{year}{1990}.
\newblock \bibinfo{title}{Integrated architectures for learning, planning, and reacting based on approximating dynamic programming}, in: \bibinfo{booktitle}{Machine learning proceedings 1990}. \bibinfo{publisher}{Elsevier}, pp. \bibinfo{pages}{216--224}.
\bibitem[{Taniguchi(2024)}]{taniguchi2024collective}
\bibinfo{author}{Taniguchi, T.}, \bibinfo{year}{2024}.
\newblock \bibinfo{title}{Collective predictive coding hypothesis: Symbol emergence as decentralized bayesian inference}.
\newblock \bibinfo{journal}{Frontiers in Robotics and AI} \bibinfo{volume}{11}.
\bibitem[{Taniguchi et~al.(2025a)Taniguchi, Hirai, Suzuki, Murata, Horii and Tanaka}]{taniguchi2025system}
\bibinfo{author}{Taniguchi, T.}, \bibinfo{author}{Hirai, Y.}, \bibinfo{author}{Suzuki, M.}, \bibinfo{author}{Murata, S.}, \bibinfo{author}{Horii, T.}, \bibinfo{author}{Tanaka, K.}, \bibinfo{year}{2025}a.
\newblock \bibinfo{title}{System 0/1/2/3: Quad-process theory for multi-timescale embodied collective cognitive systems}.
\newblock \bibinfo{journal}{arXiv preprint arXiv:2503.06138} .
\bibitem[{Taniguchi et~al.(2023a)Taniguchi, Murata, Suzuki, Ognibene, Lanillos, Ugur, Jamone, Nakamura, Ciria, Lara and Pezzulo}]{taniguchi2023world}
\bibinfo{author}{Taniguchi, T.}, \bibinfo{author}{Murata, S.}, \bibinfo{author}{Suzuki, M.}, \bibinfo{author}{Ognibene, D.}, \bibinfo{author}{Lanillos, P.}, \bibinfo{author}{Ugur, E.}, \bibinfo{author}{Jamone, L.}, \bibinfo{author}{Nakamura, T.}, \bibinfo{author}{Ciria, A.}, \bibinfo{author}{Lara, B.}, \bibinfo{author}{Pezzulo, G.}, \bibinfo{year}{2023}a.
\newblock \bibinfo{title}{World models and predictive coding for cognitive and developmental robotics: frontiers and challenges}.
\newblock \bibinfo{journal}{Advanced Robotics} \bibinfo{volume}{37}, \bibinfo{pages}{780--806}.
\newblock \URLprefix \url{https://doi.org/10.1080/01691864.2023.2225232}, \DOIprefix\doi{10.1080/01691864.2023.2225232}, \href{http://arxiv.org/abs/https://doi.org/10.1080/01691864.2023.2225232}{{\tt arXiv:https://doi.org/10.1080/01691864.2023.2225232}}.
\bibitem[{Taniguchi et~al.(2016)Taniguchi, Nagai, Nakamura, Iwahashi, Ogata and Asoh}]{taniguchi2016symbol}
\bibinfo{author}{Taniguchi, T.}, \bibinfo{author}{Nagai, T.}, \bibinfo{author}{Nakamura, T.}, \bibinfo{author}{Iwahashi, N.}, \bibinfo{author}{Ogata, T.}, \bibinfo{author}{Asoh, H.}, \bibinfo{year}{2016}.
\newblock \bibinfo{title}{Symbol emergence in robotics: a survey}.
\newblock \bibinfo{journal}{Advanced Robotics} \bibinfo{volume}{30}, \bibinfo{pages}{706--728}.
\bibitem[{Taniguchi et~al.(2024)Taniguchi, Oizumi, Saji, Horii and Tsuchiya}]{taniguchi2024constructive}
\bibinfo{author}{Taniguchi, T.}, \bibinfo{author}{Oizumi, M.}, \bibinfo{author}{Saji, N.}, \bibinfo{author}{Horii, T.}, \bibinfo{author}{Tsuchiya, N.}, \bibinfo{year}{2024}.
\newblock \bibinfo{title}{Constructive approach to bidirectional causation between qualia structure and language emergence}.
\newblock \bibinfo{journal}{arXiv preprint arXiv:2409.09413} .
\bibitem[{Taniguchi et~al.(2025b)Taniguchi, Takagi, Otsuka, Hayashi and Hamada}]{taniguchi2024cpcms}
\bibinfo{author}{Taniguchi, T.}, \bibinfo{author}{Takagi, S.}, \bibinfo{author}{Otsuka, J.}, \bibinfo{author}{Hayashi, Y.}, \bibinfo{author}{Hamada, H.T.}, \bibinfo{year}{2025}b.
\newblock \bibinfo{title}{Collective predictive coding as model of science: Formalizing scientific activities towards generative science}.
\newblock \bibinfo{journal}{Royal Society Open Science} \bibinfo{volume}{12}, \bibinfo{pages}{241678}.
\bibitem[{Taniguchi et~al.(2018)Taniguchi, Ugur, Hoffmann, Jamone, Nagai, Rosman, Matsuka, Iwahashi, Oztop, Piater et~al.}]{taniguchi2018symbol}
\bibinfo{author}{Taniguchi, T.}, \bibinfo{author}{Ugur, E.}, \bibinfo{author}{Hoffmann, M.}, \bibinfo{author}{Jamone, L.}, \bibinfo{author}{Nagai, T.}, \bibinfo{author}{Rosman, B.}, \bibinfo{author}{Matsuka, T.}, \bibinfo{author}{Iwahashi, N.}, \bibinfo{author}{Oztop, E.}, \bibinfo{author}{Piater, J.}, et~al., \bibinfo{year}{2018}.
\newblock \bibinfo{title}{Symbol emergence in cognitive developmental systems: a survey}.
\newblock \bibinfo{journal}{IEEE Transactions on Cognitive and Developmental Systems} \bibinfo{volume}{11}, \bibinfo{pages}{494--516}.
\bibitem[{Taniguchi et~al.(2023b)Taniguchi, Yoshida, Matsui, Hoang, Taniguchi and Hagiwara}]{taniguchi2022emergent}
\bibinfo{author}{Taniguchi, T.}, \bibinfo{author}{Yoshida, Y.}, \bibinfo{author}{Matsui, Y.}, \bibinfo{author}{Hoang, N.L.}, \bibinfo{author}{Taniguchi, A.}, \bibinfo{author}{Hagiwara, Y.}, \bibinfo{year}{2023}b.
\newblock \bibinfo{title}{Emergent communication through {Metropolis-Hastings} naming game with deep generative models}.
\newblock \bibinfo{journal}{Advanced Robotics} \bibinfo{volume}{37}, \bibinfo{pages}{1266--1282}.
\newblock \URLprefix \url{https://doi.org/10.1080/01691864.2023.2260856}, \DOIprefix\doi{10.1080/01691864.2023.2260856}, \href{http://arxiv.org/abs/https://doi.org/10.1080/01691864.2023.2260856}{{\tt arXiv:https://doi.org/10.1080/01691864.2023.2260856}}.
\bibitem[{Thomas and Saad(2023)}]{thomas2023neuro}
\bibinfo{author}{Thomas, C.K.}, \bibinfo{author}{Saad, W.}, \bibinfo{year}{2023}.
\newblock \bibinfo{title}{Neuro-symbolic causal reasoning meets signaling game for emergent semantic communications}.
\newblock \bibinfo{journal}{IEEE Transactions on Wireless Communications} \bibinfo{volume}{22}, \bibinfo{pages}{9101--9116}.
\newblock \DOIprefix\doi{10.1109/TWC.2023.3283392}.
\bibitem[{Tomasello(2005)}]{tomasello2005constructing}
\bibinfo{author}{Tomasello, M.}, \bibinfo{year}{2005}.
\newblock \bibinfo{title}{Constructing a language: A usage-based theory of language acquisition}.
\newblock \bibinfo{publisher}{Harvard university press}.
\bibitem[{Tucker et~al.(2022)Tucker, Levy, Shah and Zaslavsky}]{TuckerLSZ-2022-trading}
\bibinfo{author}{Tucker, M.}, \bibinfo{author}{Levy, R.P.}, \bibinfo{author}{Shah, J.}, \bibinfo{author}{Zaslavsky, N.}, \bibinfo{year}{2022}.
\newblock \bibinfo{title}{Trading off utility, informativeness, and complexity in emergent communication}, in: \bibinfo{booktitle}{Advances in Neural Information Processing Systems}.
\newblock \URLprefix \url{https://openreview.net/forum?id=O5arhQvBdH}.
\bibitem[{Ueda(2024)}]{ueda2024reinterpreting}
\bibinfo{author}{Ueda, R.}, \bibinfo{year}{2024}.
\newblock \bibinfo{title}{Reinterpreting signaling and referential games as generative models}, in: \bibinfo{booktitle}{Language Gamification - NeurIPS 2024 Workshop}.
\newblock \URLprefix \url{https://openreview.net/forum?id=6dzojDiJpc}.
\bibitem[{Ueda et~al.(2023)Ueda, Ishii and Miyao}]{UedaIM-2023-HAS}
\bibinfo{author}{Ueda, R.}, \bibinfo{author}{Ishii, T.}, \bibinfo{author}{Miyao, Y.}, \bibinfo{year}{2023}.
\newblock \bibinfo{title}{On the word boundaries of emergent languages based on harris's articulation scheme}, in: \bibinfo{booktitle}{The Eleventh International Conference on Learning Representations (ICLR), {ICLR} 2023, Kigali, Rwanda, May 1-5, 2023}, \bibinfo{publisher}{OpenReview.net}.
\newblock \URLprefix \url{https://openreview.net/pdf?id=b4t9\_XASt6G}.
\bibitem[{Ueda and Taniguchi(2024)}]{UedaT-2024-signaling-game-as-vae}
\bibinfo{author}{Ueda, R.}, \bibinfo{author}{Taniguchi, T.}, \bibinfo{year}{2024}.
\newblock \bibinfo{title}{Lewis's signaling game as beta-vae for natural word lengths and segments}, in: \bibinfo{booktitle}{The Twelfth International Conference on Learning Representations (ICLR), {ICLR} 2024, Vienna, Austria, May 7-11, 2024}, \bibinfo{publisher}{OpenReview.net}.
\newblock \URLprefix \url{https://openreview.net/forum?id=HC0msxE3sf}.
\bibitem[{Ueda and Washio(2021)}]{UedaW-2021-ZLA}
\bibinfo{author}{Ueda, R.}, \bibinfo{author}{Washio, K.}, \bibinfo{year}{2021}.
\newblock \bibinfo{title}{On the relationship between {Zipf's} law of abbreviation and interfering noise in emergent languages}, in: \bibinfo{booktitle}{Proceedings of the {ACL-IJCNLP} 2021 Student Research Workshop, {ACL} 2021, Online, JUli 5-10, 2021}, \bibinfo{publisher}{Association for Computational Linguistics}. pp. \bibinfo{pages}{60--70}.
\newblock \URLprefix \url{https://doi.org/10.18653/v1/2021.acl-srw.6}, \DOIprefix\doi{10.18653/v1/2021.acl-srw.6}.
\bibitem[{Venugopalan et~al.(2015)Venugopalan, Rohrbach, Donahue, Mooney, Darrell and Saenko}]{venugopalan2015sequence}
\bibinfo{author}{Venugopalan, S.}, \bibinfo{author}{Rohrbach, M.}, \bibinfo{author}{Donahue, J.}, \bibinfo{author}{Mooney, R.}, \bibinfo{author}{Darrell, T.}, \bibinfo{author}{Saenko, K.}, \bibinfo{year}{2015}.
\newblock \bibinfo{title}{Sequence to sequence--video to text}, in: \bibinfo{booktitle}{Proceedings of the IEEE international conference on computer vision}, pp. \bibinfo{pages}{4534--4542}.
\bibitem[{Verma and Pilanci(2025)}]{verma2025implicitly}
\bibinfo{author}{Verma, P.}, \bibinfo{author}{Pilanci, M.}, \bibinfo{year}{2025}.
\newblock \bibinfo{title}{Large language models implicitly learn to see and hear just by reading}.
\newblock \href{http://arxiv.org/abs/2505.17091}{{\tt arXiv:2505.17091}}.
\bibitem[{Vinyals et~al.(2015)Vinyals, Toshev, Bengio and Erhan}]{vinyals2015show}
\bibinfo{author}{Vinyals, O.}, \bibinfo{author}{Toshev, A.}, \bibinfo{author}{Bengio, S.}, \bibinfo{author}{Erhan, D.}, \bibinfo{year}{2015}.
\newblock \bibinfo{title}{Show and tell: A neural image caption generator}, in: \bibinfo{booktitle}{Proceedings of the IEEE conference on computer vision and pattern recognition}, pp. \bibinfo{pages}{3156--3164}.
\bibitem[{Von~Uexk{\"u}ll(1992)}]{Uexkull}
\bibinfo{author}{Von~Uexk{\"u}ll, J.}, \bibinfo{year}{1992}.
\newblock \bibinfo{title}{A stroll through the worlds of animals and men: A picture book of invisible worlds}.
\newblock \bibinfo{journal}{Semiotica} \bibinfo{volume}{89}, \bibinfo{pages}{319--391}.
\bibitem[{Wang et~al.(2024)Wang, Todd, Xiao, Yuan, C{\^o}t{\'e}, Clark and Jansen}]{wang2024can}
\bibinfo{author}{Wang, R.}, \bibinfo{author}{Todd, G.}, \bibinfo{author}{Xiao, Z.}, \bibinfo{author}{Yuan, X.}, \bibinfo{author}{C{\^o}t{\'e}, M.A.}, \bibinfo{author}{Clark, P.}, \bibinfo{author}{Jansen, P.}, \bibinfo{year}{2024}.
\newblock \bibinfo{title}{Can language models serve as text-based world simulators?}, in: \bibinfo{booktitle}{Proceedings of the 62nd Annual Meeting of the Association for Computational Linguistics (Volume 2: Short Papers)}.
\bibitem[{Wittgenstein(2009)}]{wittgenstein2009philosophical}
\bibinfo{author}{Wittgenstein, L.}, \bibinfo{year}{2009}.
\newblock \bibinfo{title}{{Philosophical Investigations}}.
\newblock \bibinfo{publisher}{John Wiley \& Sons}.
\bibitem[{Xie et~al.(2021)Xie, Qin, Li and Juang}]{xie2021deep}
\bibinfo{author}{Xie, H.}, \bibinfo{author}{Qin, Z.}, \bibinfo{author}{Li, G.Y.}, \bibinfo{author}{Juang, B.H.}, \bibinfo{year}{2021}.
\newblock \bibinfo{title}{Deep learning enabled semantic communication systems}.
\newblock \bibinfo{journal}{IEEE Transactions on Signal Processing} \bibinfo{volume}{69}, \bibinfo{pages}{2663--2675}.
\bibitem[{Xu et~al.(2015)Xu, Ba, Kiros, Cho, Courville, Salakhudinov, Zemel and Bengio}]{xu2015show}
\bibinfo{author}{Xu, K.}, \bibinfo{author}{Ba, J.}, \bibinfo{author}{Kiros, R.}, \bibinfo{author}{Cho, K.}, \bibinfo{author}{Courville, A.}, \bibinfo{author}{Salakhudinov, R.}, \bibinfo{author}{Zemel, R.}, \bibinfo{author}{Bengio, Y.}, \bibinfo{year}{2015}.
\newblock \bibinfo{title}{Show, attend and tell: Neural image caption generation with visual attention}, in: \bibinfo{booktitle}{International conference on machine learning}, pp. \bibinfo{pages}{2048--2057}.
\bibitem[{Yan et~al.(2021)Yan, Zhang, Abbeel and Srinivas}]{yan2021videogpt}
\bibinfo{author}{Yan, W.}, \bibinfo{author}{Zhang, Y.}, \bibinfo{author}{Abbeel, P.}, \bibinfo{author}{Srinivas, A.}, \bibinfo{year}{2021}.
\newblock \bibinfo{title}{Videogpt: Video generation using vq-vae and transformers}.
\newblock \bibinfo{journal}{arXiv preprint arXiv:2104.10157} .
\bibitem[{Yoshida and Taniguchi(2025)}]{yoshida2025reward}
\bibinfo{author}{Yoshida, N.}, \bibinfo{author}{Taniguchi, T.}, \bibinfo{year}{2025}.
\newblock \bibinfo{title}{Reward-independent messaging for decentralized multi-agent reinforcement learning}.
\newblock \bibinfo{journal}{arXiv preprint arXiv:2505.21985} .
\bibitem[{Yoshida et~al.(2025)Yoshida, Masumori and Ikegami}]{yoshida2025text}
\bibinfo{author}{Yoshida, T.}, \bibinfo{author}{Masumori, A.}, \bibinfo{author}{Ikegami, T.}, \bibinfo{year}{2025}.
\newblock \bibinfo{title}{From text to motion: grounding gpt-4 in a humanoid robot “{A}lter3”}.
\newblock \bibinfo{journal}{Frontiers in Robotics and AI} \bibinfo{volume}{12}, \bibinfo{pages}{1581110}.
\bibitem[{You et~al.(2024)You, Ebara, Nakamura, Taniguchi and Taniguchi}]{you2024multimodal}
\bibinfo{author}{You, Z.}, \bibinfo{author}{Ebara, H.}, \bibinfo{author}{Nakamura, T.}, \bibinfo{author}{Taniguchi, A.}, \bibinfo{author}{Taniguchi, T.}, \bibinfo{year}{2024}.
\newblock \bibinfo{title}{Multimodal continuous symbol emergence using a probabilistic generative model based on gaussian processes}, in: \bibinfo{booktitle}{2024 IEEE International Conference on Development and Learning (ICDL)}, \bibinfo{organization}{IEEE}. pp. \bibinfo{pages}{1--6}.
\bibitem[{Zaslavsky et~al.(2018)Zaslavsky, Kemp, Regier and Tishby}]{ZaslavskyKRT-2018-efficient}
\bibinfo{author}{Zaslavsky, N.}, \bibinfo{author}{Kemp, C.}, \bibinfo{author}{Regier, T.}, \bibinfo{author}{Tishby, N.}, \bibinfo{year}{2018}.
\newblock \bibinfo{title}{Efficient compression in color naming and its evolution}.
\newblock \bibinfo{journal}{Proc. Natl. Acad. Sci. {USA}} \bibinfo{volume}{115}, \bibinfo{pages}{7937--7942}.
\newblock \URLprefix \url{https://doi.org/10.1073/pnas.1800521115}, \DOIprefix\doi{10.1073/PNAS.1800521115}.
\bibitem[{Zhen et~al.(2024)Zhen, Qiu, Chen, Yang, Yan, Du, Hong and Gan}]{zhen20243d}
\bibinfo{author}{Zhen, H.}, \bibinfo{author}{Qiu, X.}, \bibinfo{author}{Chen, P.}, \bibinfo{author}{Yang, J.}, \bibinfo{author}{Yan, X.}, \bibinfo{author}{Du, Y.}, \bibinfo{author}{Hong, Y.}, \bibinfo{author}{Gan, C.}, \bibinfo{year}{2024}.
\newblock \bibinfo{title}{{3D-VLA}: A 3d vision-language-action generative world model}.
\newblock \bibinfo{journal}{Proceedings of the 41 st International Conference on Machine Learning} .
\bibitem[{Zhu et~al.(2024)Zhu, Dastani and Wang}]{zhu2024survey}
\bibinfo{author}{Zhu, C.}, \bibinfo{author}{Dastani, M.}, \bibinfo{author}{Wang, S.}, \bibinfo{year}{2024}.
\newblock \bibinfo{title}{A survey of multi-agent deep reinforcement learning with communication}.
\newblock \bibinfo{journal}{Autonomous Agents and Multi-Agent Systems} \bibinfo{volume}{38}, \bibinfo{pages}{4}.

\end{thebibliography}

\section*{Acknowledgement}
This work was supported by JSPS KAKENHI Grant Numbers JP21H04904, JP23H04835, JP23H04974, JP23K28181, and the JST Moonshot R\&D Grant Number JPMJMS2011.

\end{document}